\documentclass[final]{clv2025}

\jvol{1}
\jnum{1}
\jyear{2025}

\usepackage{xcolor}

\dochead{Survey} 

\pageonefooter{Submission received: 20 May 2025; revised version received: 09 August 2025; accepted for publication: 08 September 2025. This is a preprint of an upcoming MIT Press article.}

\usepackage{amsmath}
\usepackage{booktabs}

\runningtitle{The Quest for the Right Mediator}
\runningauthor{Mueller et al.}

\begin{document}

\title{The Quest for the Right Mediator: Surveying Mechanistic Interpretability Through the Lens of Causal Mediation Analysis}

\author{Aaron Mueller\thanks{Corresponding author. Contact at \texttt{amueller@bu.edu}.}$^{,1}$, Jannik Brinkmann$^{2}$, Millicent Li$^{3}$, Samuel Marks$^{4}$, Koyena Pal$^{3}$, Nikhil Prakash$^{3}$, Can Rager$^{3}$, Aruna Sankaranarayanan$^{5}$, Arnab Sen Sharma$^{3}$, Jiuding Sun$^{6}$, Eric Todd$^{3}$, David Bau$^{3}$,\\Yonatan Belinkov$^{7}$}

\affilblock{
    \affil{Boston University\\}
    \affil{University of Mannheim}
    \affil{Northeastern University}
    \affil{Anthropic}
    \affil{Massachusetts Institute of Technology}
    \affil{Stanford University}
    \affil{Technion -- Israel Institute of Technology}
}

\maketitle

\begin{abstract}
Interpretability provides a toolset for understanding how and why neural networks behave in certain ways. However, there is little unity in the field: most studies employ ad-hoc evaluations and do not share theoretical foundations, making it difficult to measure progress and compare the pros and cons of different techniques. Furthermore, while mechanistic understanding is frequently discussed, the basic causal units underlying these mechanisms are often not explicitly defined. In this article, we propose a perspective on interpretability research grounded in causal mediation analysis. Specifically, we describe the history and current state of interpretability taxonomized according to the types of causal units (mediators) employed, as well as methods used to search over mediators. We discuss the pros and cons of each mediator, providing insights as to when particular kinds of mediators and search methods are most appropriate. We argue that this framing yields a more cohesive narrative of the field and helps researchers select appropriate methods based on their research objective. Our analysis yields actionable recommendations for future work, including the discovery of new mediators and the development of standardized evaluations tailored to these goals.
\end{abstract}

\section{Introduction}
To understand how a neural network (NN) will generalize, we must understand the causes of its behavior. These causes include inputs, but also the intermediate computations of the network; this survey is concerned with understanding these intermediate computations. How can we understand what a NN's computations represent, such that we can obtain a deeper algorithmic understanding of \emph{how} and \emph{why} models behave the way they do?
For example, if a language model (LM) decides to refuse a user's request, was the refusal mediated by an underlying concept of toxicity, by the presence of superficial correlates of toxicity (such as the mention of particular demographic groups), or some other unexpected variable? The former would be significantly more likely to robustly and safely generalize. These questions motivate the field of mechanistic interpretability (MI),
which aims to understand how NNs arrive at particular behaviors by understanding the functional roles of their components.

\begin{figure}
    \centering
    \includegraphics[width=0.98\linewidth]{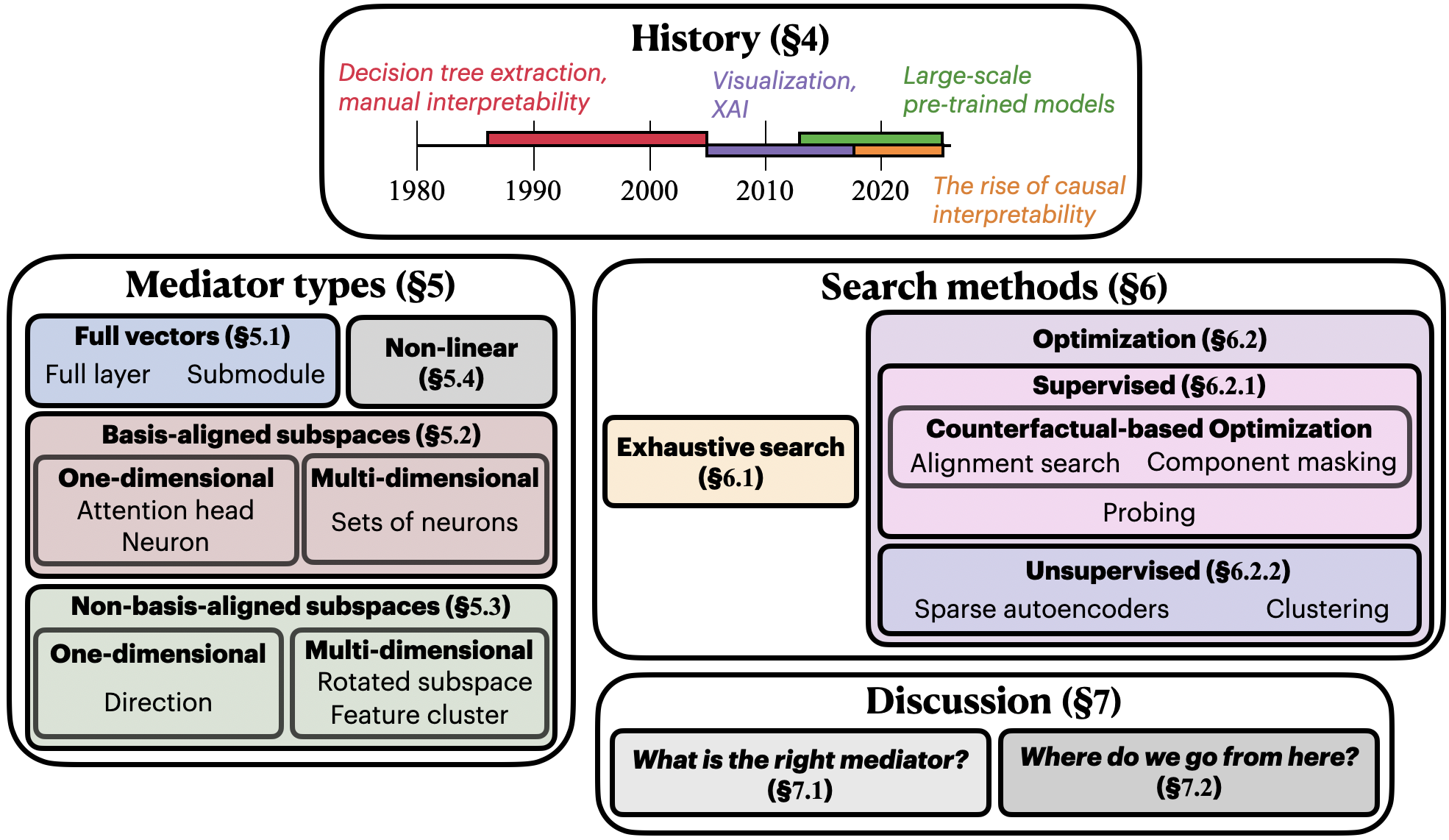}
    \caption{Outline of survey.
    We first define necessary causal terminology (\S\ref{sec:preliminaries}) and contextualize our perspective with others' (\S\ref{sec:related_work}). We then give an overview of the history of mechanistic interpretability centered on units of causal analysis (\S\ref{sec:history}). We then survey and categorize commonly used units of analysis and describe their strengths and weaknesses (\S\ref{sec:mediator_type}), as well as methods for searching over them (\S\ref{sec:mediator_search}). Finally, we discuss (\S\ref{sec:discussion}) what we consider to be among the most important questions in mechanistic interpretability: \textbf{What are the right causal abstractions for understanding and discussing the inner workings of NNs} (\S\ref{sec:right_mediator})? \textbf{What kinds of mediators and research will be needed to advance the field} (\S\ref{sec:future_work})?}
    \label{fig:overview}
\end{figure}

We view mechanistic interpretability\footnote{The meaning of ``mechanistic interpretability'' is debated; see \citet{saphra-wiegreffe-2024-mechanistic}. We define the term as any study that aims to understand, explain, or modify a neural network's behavior by studying the model's internal components, such as its representations or weights.} as equivalent to extracting \textbf{causal graphs} explaining how intermediate NN computations mediate model outputs. This framing based in causality enables a new perspective of the field: prior surveys have organized the field according to methodological differences, whereas we taxonomize work in the field according to the kinds of causal mediators---or types of nodes in the causal graphs---that a study employs (e.g., neurons, non-basis-aligned directions, attention heads, and the like). We start by describing causal mediation analysis and its role in MI (\S\ref{sec:preliminaries}); we also define the goals of MI, and goal-specific criteria by which the success of a mediator can be measured (\S\ref{ssec:criteria}). We then contrast this survey with other MI surveys (\S\ref{sec:related_work}), noting in particular the lack of surveys that center the mediator type. Following this, we present a history of mechanistic interpretability for neural networks more broadly (\S\ref{sec:history}), from backpropagation to the beginning of the current wave of mechanistic interpretability research.

We survey common mediators (units of causal analysis) used in mechanistic interpretability studies (\S\ref{sec:mediator_type}), discussing the pros and cons of each mediator type. Should one analyze individual neurons? Combinations of neurons? Full activation vectors? More broadly, \textit{what is the right unit of abstraction for analyzing and discussing neural network behaviors?} Any model component has pros and cons related to its level of granularity, whether it is a causal bottleneck, and whether it is natively part of the model (as opposed to whether it is learned via a separate module).
The mediator type determines the kinds of methods that may be used to search over them; these search methods have their own pros and cons, which we use to organize the field in \S\ref{sec:mediator_search}.

Finally, after surveying the field, we discuss practical considerations and implications for future work (\S\ref{sec:discussion}). We point out mediators which have been underexplored, but have significant potential to yield new insights; propose future mediators that are likely satisfy the criteria laid out in \S\ref{ssec:criteria}; and suggest ways to measure progress in mechanistic interpretability moving forward. Figure~\ref{fig:overview} summarizes the content of this survey.\footnote{Note that the methods discussed in this survey focus primarily on language models, so we will often use ``language model'' in place of ``neural network'' throughout this survey. However, most of the methods we discuss generalize beyond language models: most can, in theory, apply to any neural network.}

\section{Preliminaries}\label{sec:preliminaries}
\paragraph{The counterfactual theory of causality}
\citet{lewis-1973-counterfactuals} poses that a \textbf{causal dependence} holds iff the following condition holds:
\begin{quote}
``An event $E$ \emph{causally depends} on $C$ [iff] (i) if $C$ had occurred, then $E$ would have occurred, and (ii) if $C$ had not occurred, then $E$ would not have occurred.''
\end{quote}
\citet{lewis-1986-causation} extends the definition of causal dependence to be whether there is a \textbf{causal chain} linking $C$ to $E$; a causal chain is a connected series of causes and effects that proceeds from an initial event to a final one, with potentially many intermediate events between them. This idea was later extended from a binary notion of whether the effect happens at all to a more nuanced notion of causes having influence on \emph{how} or \emph{when} events occur \citep{lewis2000influence}. Other work defines notions of cause and effect as measurable quantities \citep{pearl2000causality}; this includes direct and \textbf{indirect effects} \citep{robins1992indirect,pearl2001effects}, which are common metrics in causal interpretability studies.

\begin{table}
    \centering
    \resizebox{0.75\linewidth}{!}{
    \begin{tabular}{lp{9cm}}
    \toprule
    $X$ & The input to a neural network or an exogenous variable in a causal graph. A specific value of $X$ is denoted $x$. \\
    $Y$ & The output of a neural network or outcome node in a causal graph. A specific value of $Y$ is denoted $y$. \\
    \midrule
    $\mathcal{C}$ & The computation graph of a neural network. Also used to refer to the neural network itself. \\
    $Z$ & A generic placeholder referring to any possible representation in a neural network between $X$ and $Y$. \\
    $\ell$ & A layer of $\mathcal{C}$. \\
    $\mathbf{h}^\ell$ & The representation vector at the output of layer $\ell$. \\
    $\mathbf{h}^{\ell\text{-MLP}}$, $\mathbf{h}^{\ell\text{-Attn}}$ & The vector output of the MLP or attention block, respectively, at layer $\ell$. \\
    $\mathbf{h}^\ell_i$ & A neuron in $\mathbf{h}^\ell$. \\
    $h^\ell_i$ & A scalar activation of neuron $\mathbf{h}^\ell_i$. \\
    $A^\ell_i$ & An attention head in layer $\ell$. \\
    $\mathbf{a}^\ell_i$ & The vector output (attention score, equivalent to $Q\cdot K$ before the softmax) of $A^\ell_i$. \\
    $d$ & The size of $\mathbf{h}^\ell$. \\
    \midrule
    $\mathcal{H}$ & A causal graph. \\
    $V$ & The set of nodes in causal graph $\mathcal{H}$ between $X$ and $Y$. \\
    $V_i$ & A node in $V$. A specific value of $V_i$ is denoted as $v_i$. \\
    $E$ & The set of edges in causal graph $\mathcal{H}$. \\
    $E_{i,j}$ & An edge in $E$ drawn from $V_i$ to $V_j$. \\
    \bottomrule
    \end{tabular}}
    \caption{Table of notation grouped by whether the terms refer to concepts in neural networks, causal graphs, or both.}
    \label{tab:notation}
\end{table}

\paragraph{Causal abstractions in mechanistic interpretability}
\textbf{Causal graphs} are fundamental abstractions in the causality literature \citep{pearl2000causality}. A causal graph $\mathcal{H}$ is a directed acyclic graph consisting of nodes $V$ and directed edges $E$. A \textbf{node} $V_i\in V$ corresponds to an action or event; in neural networks, it can correspond to any component (or combination thereof), as described below and in \S\ref{sec:mediator_type}. An \textbf{edge} $E_{i,j}\in E$ encodes a causal relationship between nodes, where the source is the \textbf{cause} and destination is the \textbf{effect}.\footnote{In interpretability studies, it is also frequently required that edge $E_i$ have strong influence on the final target behavior or output $Y$, rather than just the downstream intermediate component $V_j$.} For example, if edge $E_{i,j}$ is drawn from one neuron $V_i$ to another $V_j$, this indicates that $V_i$ has significant counterfactual influence over $V_j$.
Note that in a causal chain (a connected path in a causal graph), any node $V_j$ can simultaneously function as both a cause of some downstream node $V_k$ and an effect of a prior node $V_i$.

The abstraction of causal graphs extends naturally to neural networks: the computation graph of a model $\mathcal{C}$ is, by definition, the full causal graph that explains how inputs $X$ (an exogenous variable in the causal graph) are transformed into a probability distribution over outputs $Y$ (an outcome or leaf node in the causal graph). A causal node can correspond to any unit or intermediate representation $Z$ produced by the network---for example, a neuron, a full layer, an attention head, or even some grouping of these. An edge encodes a causal relationship between any two nodes in the network, where the only restriction is that the source of the edge come before the destination in the computation graph. We summarize the notation we use for describing (components of) causal graphs and computation graphs in Table~\ref{tab:notation}.

Each node can be viewed as a causal mediator that has some functional role in explaining how $X$ is transformed into $Y$. The main challenge of mechanistic interpretability studies, then, is to define a mapping from components $Z$ in computation graph $\mathcal{C}$ to a high-level causal graph $\mathcal{H}$ consisting of nodes and edges $(V,E)$ that explains how the model performs some specific behavior. This entails deciding which of the components $\{Z_i\}_{i=1}^N$ should be filtered out from the nodes of the causal graph $V$, and optionally filtering out edges between these nodes.\footnote{For some studies, it is sufficient to discover unordered sets of causally relevant components. In such cases, we assume that the causal graph is fully connected (where it is possible given the computation graph's directionality).} This survey focuses on how the type of component $Z\in \mathcal{C}$ will affect one's findings; this is discussed in detail in \S\ref{sec:mediator_type}.

In the causality literature, a \textbf{mechanism} is defined as a causal chain from cause $X$ to effect $Y$ \citep{salmon-1984-explanation,pearl-2009-causality}. The mechanistic interpretability literature, while closely related to causal interpretability,\footnote{To be more precise, we will classify any study that aims to understand a model via understanding the roles of its components or its inner representations as mechanistic interpretability. We will classify any study that employs counterfactual interventions to a model's inputs and/or representations (resulting in states the model would not naturally have taken) as causal interpretability.} does not enforce this causally grounded definition of mechanism \citep[cf.][]{miller2024transformercircuitfaithfulnessmetrics,nanda-etal-2023-emergent}. The overlap between mechanistic and causal interpretability is significant, but not total: for example, sparse autoencoders \citep{bricken2023monosemanticity,cunningham2024sparse} are correlational, but are common in mechanistic interpretability, as they can reveal the concepts encoded in a model component without requiring the researcher to hypothesize what these concepts are ahead of time. Meanwhile, methods like LIME \citep{ribeiro-2016-lime} involve interventions to input variables, but not the internals of a model.
We believe that the causal definition of ``mechanism'' is an actionable one that makes the main challenge of mechanistic interpretability more precise---to reverse-engineer algorithmic understanding or control of model behaviors, where ``algorithm'' is equivalent to a task-specific causal graph $\mathcal{H}$ explaining how the model $\mathcal{C}$ performs a given task.

\paragraph{Counterfactual interventions} In interpretability, ``causal method'' generally refers to a method that employs \textbf{counterfactual interventions} \citep{lewis-1973-counterfactuals} to some part of the model or its inputs. Early interpretability work focused on interpreting model decision boundaries by intervening on the inputs $X$ to the network \citep[e.g.,][]{ribeiro-2016-lime}, but contemporary work 
is primarily concerned with understanding which intermediary model components $Z$ are responsible for some behavior $Y$---i.e., finding the model components $Z\in\mathcal{C}$ from the low-level computation graph to keep as nodes $V\in\mathcal{H}$ in the high-level causal graph  \citep[e.g.,][]{geiger2021causal,hanna2023how}.

\begin{figure}[t]
    \centering
    \includegraphics[width=0.9\linewidth]{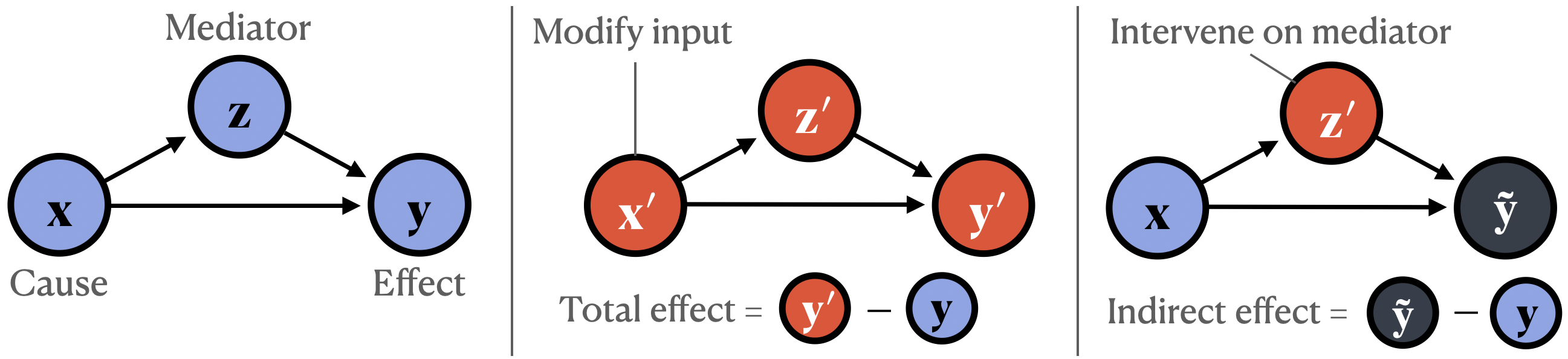}
    \caption{Visual summary of causal mediation analysis. Given input $X=\color{blue}{x}$ and the resulting output (model prediction) $Y=\color{blue}{y}$, and another input $X=\color{red}{x'}$ that results in different output $Y=\color{red}{y'}$, we can compute the total effect of changing $\color{blue}{x}$ to $\color{red}{x'}$ as $\color{red}{y'} \color{black}{-} \color{blue}{y}$. In neural networks, there exist components $Z$ that mediate the influence of $X$ on $Y$. A common way to quantify the importance of $Z$ is by measuring its \textbf{indirect effect} (Eq.~\ref{eq:ie}), where, given $X=\color{blue}{x}$, one sets $Z$ to some counterfactual value $\color{red}{z'}$. In this figure, we set $Z$ to what it would have been given $\color{red}{x'}$; this results in $Y=\color{darkgray}{\tilde{y}}$. One can then measure the indirect effect as $\color{darkgray}{\tilde{y}}\color{black}{-}\color{blue}{y}$.}
    \label{fig:causal-metrics}
\end{figure}

Causal mediation analysis \citep{pearl2001effects} provides a framework for performing counterfactual interventions and interpreting their results. Given input $X$, output $Y$, and a causal graph consisting of many intermediate nodes $V$ between $X$ and $Y$, the causal influence of an intermediate node (mediator) $V_i\in V$ on the output $Y$ is quantified as $V_i$'s \textbf{indirect effect} (IE; \citealp{pearl2001effects,robins1992indirect}). This metric is based on the notion of counterfactual dependence, where one measures the difference in some target metric $m$ before and after intervening on the mediator. In practice, $m$ is typically (but not always) a function of the model's output $Y$.\footnote{$m$ can be any scalar value, including the value of an intermediate causal variable $\in V$, or even the aggregated values of multiple intermediate variables. For example, ACDC \citep{conmy2023automated} recursively computes a circuit by first finding components with high IE on $Y$ (where $m$ is derived from $Y$), and then finding components with high IE on those components (where $m$ is now derived from intermediate components $Z$), and so on.} More precisely, one starts by measuring $m$ given a normal run of the model on input $X=x$, where $V_i$ takes its natural value(s) $v_i$. We then compare this to $m$ given $X=x$ where we intervene to set mediator $V_i$ to some alternate (counterfactual) value $v_i'$:\footnote{Appendix~A surveys methods for sourcing $v_i'$. This value can come from alternate inputs where the answer is flipped, means over many inputs, or arbitrary constants (typically $0$).}
\begin{align}
    \text{IE}(m; X, x; V_i, v_i, v_i') = m(V_i=v_i\mid X=x) - m(\text{do}(V_i = v_i')\mid X=x),
\label{eq:ie}
\end{align}
where ``do'' is an operation where the experimenter intervenes, setting a value in the experiment to one that it would not naturally have taken without the experimenter's involvement. See Figure~\ref{fig:causal-metrics} for an illustration of the indirect effect given a neural network component $Z$.\footnote{The choice of counterfactual $v_i'$ significantly affects which components will be identified as influential. Common approaches for selecting $v_i'$ include using values from alternative inputs $x'$, applying a constant (e.g., 0), or adding noise $\epsilon$, where $\epsilon$ is typically scaled according to the norm or variance of $v_i$. These are conceptually distinct operations that will uncover different kinds of components; we provide a more detailed discussion in Appendix~A. We leave this to the appendix, as this survey is concerned with how results are affected by the choice of mediator $Z$---a topic distinct from the choice of intervention value $v_i'$.}

\paragraph{From intervention to evaluation} The tools of causal mediation analysis allow us to quantify the causal influence of a mediator on model behavior. However, identifying influential components is not sufficient for mechanistic understanding on its own. Depending on the goal of a study, we may define additional criteria by which the appropriateness or quality of a mediator may be judged. In the following subsection, we outline an evaluation framework that will help frame the following survey of mediator types used in existing research.

\subsection{Evaluation Criteria}\label{ssec:criteria}
The goals of a mechanistic interpretability study determine how success should be defined. We identify three primary goals: (1) explaining model behavior, (2) verifying a mechanistic hypothesis, and (3) localization and editing. Below, we define criteria by which we can compare the utility of mediator types.

First, we note four criteria that apply to each goal. At a high level, a good causal explanation explains the most general effects from the fewest non-trivial causes.\footnote{We add the ``non-trivial'' qualifier because one could trivially maximize the scope of an explanation by taking the entire computation graph $\mathcal{C}$ as a single mediator.} Thus, the ideal causal mediator is one that achieves a Pareto optimum between maximal \textbf{sparsity} (using as little of the computation graph as possible), \textbf{generality} (explaining the widest breadth of data), \textbf{selectivity} (explaining only the phenomenon of interest and as little else as possible), and \textbf{faithfulness} (retaining fidelity to the original model).

\paragraph{Explaining model behavior} Often, our goal is to produce qualitative and human-understandable insights about how models perform a certain behavior or task. What makes a good explanation? In mechanistic interpretability, this can be defined as one that maximizes the above metrics of sparsity, generality, selectivity, and faithfulness, as well as \textbf{human understandability}. Faithfulness is already a common metric in mechanistic explanations \citep{hanna2024faithfaithfulnessgoingcircuit,marks2024sparse,wang2023interpretability}; measuring it is straightforward, as it is a generalization of causal mediation analysis to evaluation.\footnote{We refer readers to \citet{hanna2024faithfaithfulnessgoingcircuit} for further detail.} Sparsity is often called ``minimality'' \citep{mueller2025mibmechanisticinterpretabilitybenchmark,wang2023interpretability}; this metric is not as common as faithfulness, but is becoming increasingly common in circuit analyses. It is typically either defined as the inclusion of as few redundant dependencies as possible \citep{wang2023interpretability}, or as the inclusion of as few components as is necessary to achieve high faithfulness \citep{mueller2025mibmechanisticinterpretabilitybenchmark}. Generality is a rarer metric, as it requires evaluating on out-of-distribution examples that may need to be carefully curated. Recent work has argued that generality is a crucial metric, and has begun to define ways to measure it \citep{huang2025internal,li2025interpretationpredictbehaviorunseen}. Selectivity is not often explicitly discussed nor measured, as it is assumed that maximizing faithfulness and sparsity should implicitly optimize selectivity; more work is needed to verify this assumption. The most difficult of these criteria to quantify is human understandability; well-trained sparse autoencoder features are typically easier to interpret than neurons, but it is not yet clear whether this metric can be measured in a reliable manner.

\paragraph{Verifying a mechanistic hypothesis} Sometimes, we already have a well-defined guess as to how a model accomplishes a task (a \emph{mechanistic hypothesis}), and we would like to verify to what extent our hypothesis is accurate. Here, the criteria are similar as when explaining model behaviors: we would like to locate the smallest set of components or lowest-rank subspace that aligns best with the hypothesized explanation on the broadest possible data distribution. The primary difference is that the human-understandability of the mediator is of lesser importance for this goal, as understandability is a function of the human's mechanistic hypothesis and its accuracy, rather than the method used to align the model with the hypothesized causal variables. Criteria for success thus include sparsity, generality, and selectivity, as well as a modified form of faithfulness that we term \textbf{counterfactual faithfulness}. Counterfactual faithfulness measures whether the model's behavior changes in the expected manner when we perform counterfactual interventions to a specific part of the hypothesized mechanism; this is typically quantified as the interchange intervention accuracy (IIA; \citealp{geiger2021causal}).

\paragraph{Localization and editing} We often simply want to know where in a model some ability is implemented without necessarily understanding the components that implement it. This enables applications like model editing, steering, and parameter-efficient model adaptation (e.g., with LoRAs applied to specific layers). Here, we want to maximize sparsity, generality, and selectivity, but we do not assign as great an importance to human understandability, and do not necessarily require a hypothesis for this to work well.
If using localized components for an application like steering or model editing, then one should maximize primarily for \textbf{downstream task performance}, with sparsity and generality ideally being integrated into the downstream evaluation metrics. Note that high-quality evaluation is essential: one can maximize the efficacy of model editing via fine-tuning all model parameters, but this may produce unintended side effects if sparsity is ignored, and may not generalize well out-of-distribution if generality is ignored. Put simply, if there exist multiple mediator types that achieve similar test performance, then the best solution is likely that which then also attains the greatest sparsity and generality. Note that faithfulness is also important, but is not explicitly considered in model editing because it is trivially satisfied if the model's behavior changes after the editing operation.

\section{Related Work}\label{sec:related_work}
Causal interpretability surveys do not always focus on model internals, and mechanistic interpretability surveys do not necessarily require causal grounding. We give a brief overview of both types of survey here, contrasting them with ours. 
We also discuss recent tooling efforts that have accompanied the growing interest in mechanistic interpretability.

\paragraph{Mechanistic/model-internal interpretability surveys} Some surveys catalogue studies that aim to understand the latent representations of neural networks \citep{belinkov-glass-2019-analysis,belinkov2021ProbingSurvey,sajjad-etal-2022-neuron}; these have often called for more causal validations of correlational observations. More recent surveys tend to focus increasingly on giving practical overviews of how to use common methods for intervening on model internals \citep{ferrando2024primer,rai2024practical}. Others provide perspectives for understanding the trajectory of the mechanistic interpretability field \citep{räuker2023transparentaisurveyinterpreting}, and/or cataloguing the impacts of the field \citep{bereska2024mechanistic}. Notably, each of these surveys taxonomizes the field based on methodological differences; for example, a common contrast might be circuit analysis (discovering causal graphs of task-specific causal dependencies from model components, as in \citealp{wang2023interpretability,conmy2023automated}) vs.\ causal variable alignment methods (aligning model representations to human-provided concepts, as in \citealp{geiger2021causal,wu2023interpretability}), even if they both operate over the same kinds of components.

There is a gap here: many of these methods implicitly deploy the same units of analysis, and thus benefit/suffer from the same fundamental pros/cons as a result. For example, it is not clear whether one would be better off discovering circuits over neurons, or attention heads, or abstractions over groupings of these. The same issue applies to any model-internals method. Thus, in this survey, we instead foreground the units of causal analysis that a study employs, as well as the way in which the study searches over those units, as primary factors in categorizing the study. We also ground the field in the language of causality, which grounds the goals of mechanistic interpretability: to discover causal subgraphs explaining how inputs are transformed into outputs.

\paragraph{Causal interpretability surveys} 
\citet{moraffah2020causal} is a causal interpretability survey that categorizes various streams of causal interpretability research according to the methods they employ, though the studies they summarize are not necessarily based in the ideas of causal mediation analysis. The units of analysis were also not foundational to their organization, nor directly compared to each other. Other interpretability surveys \citep{subhash2022what, gilpin2018explaining, singh2024rethinking} focus on methods for explaining the decisions of neural networks without causally grounding the explanation methods or focusing on model internals. Many causality-focused surveys are domain-specific, including areas such as cybersecurity \citep{10.1145/3665494} and healthcare \citep{wu2024causal}. Some focus on particular domains; for example, in NLP, some focus on how causal inference can improve interpretability \citep{feder-etal-2022-causal}, or ways to explain \citep{danilevsky-etal-2020-survey,lyu2024toward} or interpret \citep{madsen2022posthoc} neural NLP systems. Our survey is more specifically focused on studies that aim to understand NLP systems via their internal components---and even more specifically, those that do so via causal techniques such as interventions to those components.

\paragraph{Tools} Several libraries have recently been released to facilitate causal interpretability methods that involve interventions to model components. These tools can implicitly prioritize certain types of mediators over others.
For instance, \texttt{pyvene}~\citep{wu2024pyvene} is designed specifically to aid in locating non-basis-aligned multidimensional subspaces via alignment search methods such as distributed alignment search and its successors~\citep{geiger2024finding, wu2023interpretability, huang2024ravel, wu2024reft}. While it can also be used for other kinds of model interventions, this library could be particularly useful for those wishing to verify existing causal hypotheses.
\texttt{TransformerLens}~\citep{nanda2022transformerlens} and libraries based on it \cite[\texttt{Prisma};][]{joseph2023vit} are interpretability tools for examining Transformer-based neural networks. In these libraries, the interface is standardized across model architectures. This tends to encourage a focus on basis-aligned components such as neurons and attention heads, subspaces, and layers, as interventions to these mediators are natively supported. \texttt{NeuroX} \citep{dalvi-etal-2023-neurox} similarly incentivizes neuron-level interpretability in particular.
\texttt{NNsight}~\citep{fiottokaufman2024nnsightndifdemocratizingaccess} and \texttt{Baukit} \citep{bau2022baukit} are more transparent interfaces that provide access to the underlying PyTorch model architecture, which allows for flexible modifications of the model's computation graph. Due to different naming conventions across model developers, this more transparent access may make it harder to generalize basis-aligned intervention code across architectures at first, but research on both basis-aligned and non-basis-aligned mediators is more accessible under this paradigm.

Note that this survey is intended more as a scientific review of the field rather than a practical guide to using these tools, so we have mainly discussed these toolkits with respect to the mediators that they enable working with. Some surveys such as as~\citet{ferrando2024primer,rai2024practical} or code tutorials~\citep{nanda2022transformerlens, wu2024pyvene, fiottokaufman2024nnsightndifdemocratizingaccess, mohebbi2024transformertutorial} make hands-on practical introduction to particular methods their explicit purpose, without necessarily assuming (nor discussing the benefits nor drawbacks of) a particular categorization of methods, nor their units of analysis. See these surveys for more hands-on guides to implementing interpretability methods.

\section{Lessons from the History of Interpretability}\label{sec:history}
Causal interpretability techniques have existed since the beginning of deep learning. What distinguishes the current wave of mechanistic interpretability studies from past causal interpretability work? What actionable lessons can past work (which often used very different methods and mediators to contemporary studies) teach us about analyzing intermediate model computations? We claim that the lens of causal mediation analysis (1) enables a novel and clear narrative of the trajectory of interpretability research; (2) links current issues in the field to longstanding issues that have existed since at least the 1980s; and (3) highlights actionable research directions. We focus primarily on (1) and (2) in this section, and return to (3) in \S\ref{sec:discussion}.

\paragraph{Interpretability at the beginning of deep learning} In 1986, \citeauthor{rumelhart-1986-backprop} published an algorithm for neural network backpropagation and an analysis of this algorithm. This enabled and massively popularized research into multi-layer perceptrons (MLPs)---now often called feedforward layers. That work arguably represents the first mechanistic interpretability study: the authors evaluated their method by inspecting each activation and weight in the neural network, and observing whether the learned algorithm corresponded to the human intuition of how the task should be performed. In other words, they reverse-engineered the algorithm of the network by labeling the rules encoded by each neuron and weight!

From the 1980s through the early 2000s, rule extraction via neuron-level activation and weight analysis remained popular. At first, this was a manual process: networks were either small enough to be manually interpreted \citep{rumelhart-1986-backprop,McClelland1985DistributedMA} or interpreted with the aid of carefully crafted datasets \citep{elman1989representation,Elman1990FindingSI,elman-1991-distributed}. For example, \citet{Elman1990FindingSI,elman-1991-distributed} found that recurrent neural networks were capable of capturing hierarchical semantic relationships, and were sensitive to syntactic context.
Alternatively, researchers could prune the network \citep{mozer-1988-pruning, karnin-1990-pruning} to a sufficiently small size to be manually interpretable. 
Later, researchers proposed techniques for automatically extracting rules 
\citep{hayashi-1990-rules} or decision trees from NNs \citep{craven-1994-tree,craven1995extracting,krishnan-1999-decision,boz-2002-tree}---often after the network had been pruned. At this point, interest in automated causal methods based on interventions had not yet been established, as networks were often small and simple enough to directly understand without significant abstraction.

Nonetheless, as the size of neural networks increased, the number of rules that could be encoded in a network increased. Thus, rule/decision tree extraction techniques could not generate easily human-interpretable explanations nor algorithmic abstractions of model behaviors beyond a certain size.
This led to the rise of \textbf{visualization methods} in the 2000s, which became a popular way to demonstrate the complexity of phenomena that models had learned to encode. Visualizations of network inputs and outputs \citep{tzeng-2005-opening} and interactive visualizations of model activations \citep{erhan-2009-visualizing} were valuable initial tools for generating hypotheses as to what kinds of concepts models could represent.
While visualization research was generally \emph{not} causal, this subfield would remain influential for interpretability research as neural networks scaled in size in the following decade.

\paragraph{Large-scale pre-trained models} The 2010s were a time of rapid change in machine learning. In 2012, the first large-scale and widely adopted pre-trained neural network, AlexNet \citep{krizhevsky-2012-alexnet}, was released. Not long after, pre-trained word embeddings \citep{mikolov-2013-word2vec1,mikolov-2013-word2vec2,pennington-etal-2014-glove} became common in NLP, and further pre-trained deep networks followed \citep{he2016deep}. These were based on ideas from \emph{deep learning}. This represented a significant paradigm shift: formerly, each study would build ad-hoc models which were not shared across studies, but which were generally more transparent.\footnote{Many systems built before deep learning were based on feature engineering, and so the information they relied on was more transparent than in current systems.} 
After 2012, there was a transition toward using a shared collection of significantly larger and more capable---but also more opaque---models.
This raised new questions on what was encoded in the representations of these shared scientific artifacts.
The rapid scaling of these models rendered old neuron-level rule extraction methods either intractable or made its results difficult to interpret. Thus, interpretability methods in the early 2010s deployed scalable and relatively fast \emph{correlational} methods, including visualizations \citep{zeiler-2014-visualizing} and saliency maps \citep{simonyan2014deep}.
This trend continued into 2014--2015, when recurrent neural network--based \citep{Elman1990FindingSI} language models \citep{mikolov-2010-interspeech} began to overtake non-neural statistical models in performance \citep{bahdanau2015neural}; for example, visualizing RNN and LSTM \citep{hochreiter1997lstm} hidden states was proposed as a way to better understand their incremental processing \citep{karpathy-2015-visualizing,strobelt2017lstmvis}.

At the same time, interpretability methods started to focus more on \emph{explaining} model predictions.\footnote{There has classically been a distinction between \emph{local} and \emph{global} interpretability; local interpretability is concerned with explaining specific model predictions, whereas global interpretability is concerned with explaining a given model behavior in general across examples \citep{lipton2018mythos,guidotti2019explaining}.
Both styles of interpretability can be valuable, depending on one's research question. A benefit of causal mediation analysis is that it can encompass both styles. As recent work has tended to focus more on global interpretability, we devote more attention to this style of work, though we cite and acknowledge examples of local interpretability methods in this section.}
The explainable AI (XAI) field was and is extensive. One line of work designed supervised auxiliary (correlational) models to explain particular model predictions, such as LIME \citep{ribeiro-2016-lime,ribeiro2016modelagnostic}, Anchors \citep{ribeiro2018anchors}, and extensions like CLEAR that explicitly integrate notions of counterfactual fidelity to the output explanations \citep{white2020measurable}. These models learn local decision boundaries, or some human-interpretable simplified representation of a model's behavior. Other works interpreted predictions via feature importance measures like SHAP \citep{lundberg-2017-unified}. Influence functions \citep{koh2017understanding} traced the model's behavior back to specific instances from the training data. Another line of work sought to directly manipulate intermediate concepts to control model behavior at test time \citep{koh2020concept}, or to decompose distributed representations into interpretable symbolic representations post hoc \citep{odense2020layerwise}. The primary difference between these visualization-/correlation-/input-based methods and current methods lies in whether they prioritize black-box explanations or white-box explanations---that is, whether they explain model behaviors in terms of input/output relationships or require analysis of model internals, respectively. Black-box explanations allow us to generate hypotheses as to the types of \emph{input concepts} that explain particular model predictions. In contrast, current work prioritizes \emph{white-box explanations}---i.e., highly localized and causal explanations of \emph{how} and in \emph{which components of the computation graph} models perform a given behavior.

2017--2019 featured perhaps the largest architectural shift (among many) in machine learning methods at this time: Transformers \citep{vaswani2017attention} were released and quickly became popular due to scalability and high performance. This led directly to the first successful large-scale pretrained language models, such as (Ro)BERT(a) \citep{devlin-etal-2019-bert,liu2019roberta} and GPT-2 \citep{radford2019language}. These significantly outperformed prior models, but it was unclear \emph{why}---and at this scale, analyzing neural networks at the neuron level using past techniques had become intractable. This combination of high performance and little mechanistic understanding created demand for interpretability techniques that allowed us to see \emph{how} language models had learned to perform so well.

Hence, correlational probing methods rose to meet this demand.  In this approach, classifiers are trained on intermediate activations to extract some target phenomenon. Probing classifiers have been used to investigate the latent morphosyntactic structures encoded in static word embeddings \citep{kohn-2015-whats,gupta-etal-2015-distributional} or intermediate hidden representations in natural language systems---for example, in neural machine translation systems \citep{shi-etal-2016-string,belinkov-etal-2017-neural,conneau-etal-2018-cram} and pre-trained language models \citep{hewitt-manning-2019-structural,hewitt-etal-2021-conditional,lakretz-etal-2019-emergence,lakretz2021rnnslearnrecursivenested}.
However, probing classifiers lack consistent baselines, and the claims made in these studies were not often causally verified \citep{belinkov2021ProbingSurvey}.
For instance, although an intervention may target a causal property of the task $V_i \rightarrow Y$, an alternative spurious property $V_j$ may be picked up by the probe, which impedes causal claims about $V_i \rightarrow Y$ \citep{ravichander-etal-2021-probing}. This encouraged researchers to search for more causally efficacious methods.

\paragraph{The rise of causal and mechanistic interpretability} 2017--2018 featured the first hints of our current wave of mechanistic interpretability, primarily based on interventions to neurons or full layers. \citet{giulianelli-etal-2018-hood} trained a probing classifier, but then used gradients from the probe to modify the activations of the network. Other studies analyzed the functional role of individual neurons in static word embeddings \citep{li2017understandingneuralnetworksrepresentation} by forcing certain neurons on or off. Parallel developments in computer vision were influential: \citet{bau2018visualizing} found that interpretable concepts in the outputs of generative adversarial networks \citep{goodfellow-2014-gan} could be modified via interventions to specific neurons.
The idea of manipulating neurons to steer behaviors was then applied to downstream task settings, such as machine translation \citep{BauBSDDG19}. These techniques were popularized in 2020 when \citet{vig2020causal} proposed a method for assigning task-specific causal importance scores to specific neurons and attention heads by systematically computing each component's indirect effect (Eq.~\ref{eq:ie}) on the model's output. It was an application of the counterfactual theory of causality \citep{lewis-1973-counterfactuals,lewis-1986-causation}, as well as Judea Pearl's causal mediation analysis framework \citep{pearl2001effects,pearl2000causality}. 
This enabled a new line of interpretability research that aimed to faithfully localize model behaviors to specific components---an idea that would become foundational to contemporary causal and mechanistic interpretability.

At the same time, however, researchers began to realize the significant performance improvements that could be gained by massively increasing the number of parameters and training corpus sizes of neural networks \citep{brown-2020-gpt3,kaplan2020scaling}.
Increasing model sizes resulted in more interesting subjects of study, but also rendered causal interpretability significantly more difficult. Thus, a primary challenge in interpretability has been to balance the often contradictory goals of (i) obtaining a causal understanding of how and why models behave in a given manner, while also (ii) designing methods that are efficient and scalable.

Presently, there exist many subfields of interpretability that propose and apply causal methods to understand which model components contribute to an observed model behavior~\citep[e.g.,][]{elhage2021mathematical,geiger2021causal,conmy2023automated}.
There have also been efforts to discover more human-interpretable mediators by moving toward latent-space structures aside from (collections of) neurons \citep{cunningham2024sparse,bricken2023monosemanticity,wu2023interpretability}.
These methods and the units they are based on form the focus of this survey.

\paragraph{The lens of causal mediation analysis} For much of the history of deep learning, layers (\S\ref{ssec:layers-submodules}) and neurons (\S\ref{sec:neurons_heads}) were the basic unit of study in mechanistic interpretability. They are natural units of the model (i.e., require no external modules to discover), thus making them faithful to the model's computations by definition. In toy models, they can sometimes be human-interpretable, and their simplicity and small quantity enable researchers to exhaustively search over all of them (\S\ref{ssec:exhaustive-search}). Large-scale pre-trained models, however, contain far too many neurons for such methods to be tractable. Furthermore, their neurons are typically \emph{not} straightforwardly interpretable because representations in neural networks are generally \textbf{distributed} \citep{hinton1986distributed}; in other words, there is a many-to-many relationship between neurons and concepts. Thus, the field has turned to more sophisticated abstractions like \textbf{sets of neurons}, \textbf{attention heads} (\S\ref{sec:neurons_heads}), or even \textbf{non-basis-aligned subspaces} (\S\ref{sec:non_basis_subspaces}) that require external modules (such as probes or sparse autoencoders) to locate (\S\ref{ssec:continuous-search}). Each of these mediator types has strengths and weaknesses. A mediator type also determines, to a large extent, the kinds of concepts that can be found, and the class of methods that can be used to find them. In the following section, we more precisely define these units of analysis, and compare their strengths and weaknesses (\S\ref{sec:mediator_type}). Then, we give practical context to each mediator type by describing the methods that can be used to find them (\S\ref{sec:mediator_search}), and the strengths and weaknesses thereof.

\begin{figure}
    \centering
    \includegraphics[width=0.9\linewidth]{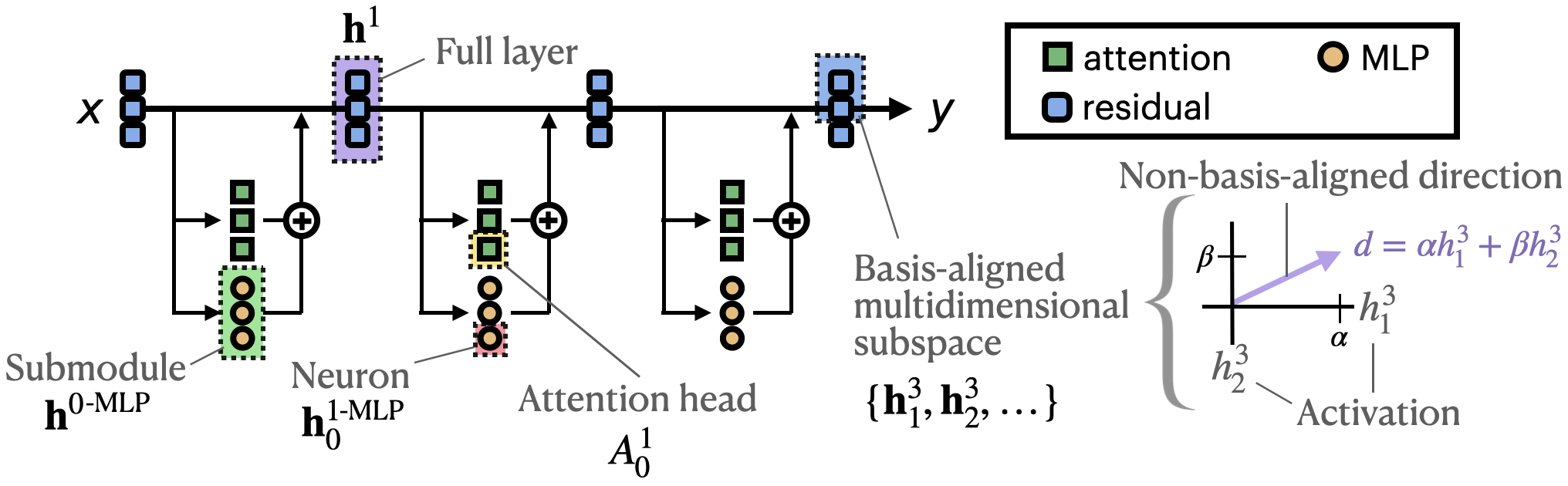}
    \caption{Visualization of common mediator types in neural networks. \textbf{Neurons} or \textbf{attention heads} are common units of analysis. \textbf{Full layer} and \textbf{submodule} vectors are more coarse-grained, but more easily enumerable. One can also implicate a \textbf{multidimensional subspace}, which could be neuron-basis-aligned (as in a group of neurons, pictured here) or non-basis-aligned. Non-basis-aligned mediators---e.g., arbitrary \textbf{directions} in activation space---have recently become a popular mediator type due to their monosemanticity. However, discovering non-basis-aligned mediators requires external modules such as classifiers, autoencoders, or other modifications to the original computation graph. Note that while this figure depicts a Transformer, many of the mediator types generalize to other architectures (the primary exception being attention heads).}
    \label{fig:compute-graph}
\end{figure}

\section{Mediator Types}\label{sec:mediator_type}
In this section, we discuss different types of causal mediators in neural networks, and the pros and cons of each. Figure~\ref{fig:compute-graph} visualizes a computation graph of a Transformer-based language model, and units in the graph that are often used as mediators in mechanistic interpretability studies. In mechanistic interpretability, we often do not want to treat the full computation graph as the final causal graph, as it is large and difficult to directly interpret. Thus, we typically want to build higher-level causal abstractions that capture only the most important mediators, and/or where each causal node is human-interpretable.

In this section, our primary questions are: What kinds of model components can be used as mediators? What are the strengths and weaknesses of using particular kinds of components as mediators? In this subsection, we  define each mediator type. Then, in the following mediator-specific subsections, we discuss how they have been used in interpretability, and their pros and cons. Table~\ref{tab:mediators} summarizes mediator types, their strengths and weaknesses, and search methods
commonly used to identify each.

\begin{table}[t]
    \caption{Summary of mediator types, the pros and cons of each, the search methods that are typically used to search over them, and examples of studies that employ them.}
    \resizebox{\linewidth}{!}{
    \begin{tabular}{p{3.45cm}p{3.5cm}p{4.8cm}p{3.75cm}p{4.55cm}}
    \toprule
    \textbf{Mediator type} & \textbf{Strengths} & \textbf{Weaknesses} & \textbf{Common search methods} & \textbf{Example studies} \\
    \midrule
    Full layers and submodules (\S\ref{ssec:layers-submodules}) & Small search space. Some useful applications. & Difficult to interpret. Not well-suited to explaining model behavior. & Exhaustive search (\S\ref{ssec:exhaustive-search}), supervised probing (\S\ref{sssec:supervised}) & \citet{hupkes2017diagnostic,giulianelli-etal-2018-hood,hewitt-manning-2019-structural,geva2023dissecting,meng2022locating} \\\midrule
    Neurons and attention heads (\S\ref{sec:neurons_heads}) & Discrete and enumerable. Relatively fine-grained; sometimes enables model control and editing. & Search space not always tractable. Often not interpretable. & Exhaustive search (\S\ref{ssec:exhaustive-search}), Optimization (\S\ref{ssec:continuous-search})& \citet{vig2020causal,bau2020understanding,finlayson-etal-2021-causal,lakretz-etal-2019-emergence,cao-etal-2021-low} \\\midrule
    Non-basis-aligned spaces (\S\ref{sec:non_basis_subspaces}) & Fine-grained. Interpretable. Enables precise control of NN behaviors. & Non-enumerable. Typically requires optimization; sensitive to training setup and random variance. May not be faithful to original model. & Optimization (\S\ref{ssec:continuous-search}): supervised probing (\S\ref{sssec:supervised}), unsupervised methods such as sparse autoencoders (\S\ref{sssec:unsupervised}) & \citet{wu2023interpretability,bricken2023monosemanticity,cunningham2024sparse,marks2023geometry,marks2024sparse,ravfogel-etal-2021-counterfactual} \\
    \bottomrule
    \end{tabular}}
    \label{tab:mediators}
\end{table}

One possible mediator type is a \textbf{full layer}---typically the output activations or \emph{hidden state} $\mathbf{h}^\ell$ of a specific layer $\ell$ (\S\ref{ssec:layers-submodules}). Each index $\mathbf{h}_i^\ell$ is a \textbf{neuron} that can take some activation $\mathbf{h}_i^\ell = h_i^\ell$.\footnote{In other words, we use ``neuron'' to refer to any basis-aligned direction in activation space.} One can also use the output vector of an intermediate \textbf{submodule} within the layer (e.g., an MLP), rather than the output of the whole layer.
For example, in Transformers~\citep{vaswani2017attention},\footnote{Transformers are currently the dominant architecture for language models; as such, most  work in this space focuses on this architecture.
However, our ideas are presented in a general way that will also apply (with minor modifications) to other neural network--based architectures, such as recurrent neural networks \citep{mikolov-2010-interspeech} and state space models~\citep{gu2022efficiently,gu2024mamba}.} a layer typically consists of two submodules: a multi-layer perceptron (MLP) and an attention block, which can be arranged either sequentially or in parallel.
The outputs of these submodules $\mathbf{h}^{\ell\text{-MLP}}$ and $\mathbf{h}^{\ell\text{-Attn}}$ are also activation vectors, so we will
refer to their individual dimensions as neurons as well.\footnote{Using the same notation emphasizes that these are mediators of the same level of granularity. However, we acknowledge that this obscures that neurons in different locations often encode different types of features.}

One can also group neurons into sets, and use a neuron set as a single mediator (\S\ref{sec:neurons_heads}). A set of neurons~(possibly of size $1$) $\{\mathbf{h}_i^\ell,\mathbf{h}_j^\ell,\ldots\}$ from a vector $\mathbf{h}^\ell$, is referred to as a \textbf{basis-aligned subspace} of $\mathbf{h^\ell}$. 
A one-dimensional basis-aligned subspace is equivalent to a neuron; for clarity, we will use basis-aligned subspace primarily to refer to multidimensional spaces (sets of neurons of size $>1$).

Basis alignment refers to whether concept representations are aligned with specific dimensions of $\mathbf{h}^\ell$ (as contrasted with weighted combinations of dimensions). Basis alignment is a key concept: if a mediator $V$ is aligned with the latent space basis vectors defined by a (group of) neuron(s) $\{\mathbf{h}^\ell_i, \mathbf{h}^\ell_j, \ldots\}$, then we can discover it via non-parametric methods. For example, it is straightforward to exhaustively search over and intervene on individual neurons; it is less tractable, but still theoretically possible, to enumerate all $2^n-1$ possible combinations of neurons without using any additional parameters. However, causally relevant mediators are not guaranteed to be aligned with neurons in activation space; indeed, recent work has found human-interpretable features in arbitrary directions that are \emph{not} aligned to neuron bases \citep{elhage2022superposition,bricken2023monosemanticity}.

Thus, in recent studies, it is common to study \textbf{non-basis-aligned spaces} (\S\ref{sec:non_basis_subspaces}). 
Each dimension
in a non-basis-aligned subspace can be defined as a weighted linear combination of neuron activations.
For example, to obtain a non-basis-aligned \textbf{direction},\footnote{We use ``direction'' to refer to one-dimensional spaces.} we could learn coefficients $\alpha$ and $\beta$ to weight the activations of neurons $\mathbf{h}_i^\ell$ and $\mathbf{h}_j^\ell$ (optionally with a bias term $\mathbf{b}$):
\begin{equation}
\mathbf{d} = \alpha \cdot \mathbf{h}_i^\ell + \beta\cdot \mathbf{h}_j^\ell + \ldots + \mathbf{b},
\label{eq:direction}
\end{equation}
where neurons $\mathbf{h}_i^\ell$ and $\mathbf{h}_j^\ell$ are allowed to take their natural activations given some input $x$, but $\alpha$ and $\beta$ remain fixed across inputs. Note that $\alpha$ and $\beta$ are \emph{not} part of the original computation graph $\mathcal{C}$. This means that discovering non-basis-aligned directions often requires external modules that weight components from the computation graph in some way---e.g., classifiers or autoencoders.

The primary trade-off between these mediator types is their granularity and quantity. This section proceeds from coarser to finer granularity and in increasing quantity. Broadly speaking, finer-grained mediators are more likely to optimize sparsity and selectivity, but are more difficult to search over, and may not be faithful to the original model if optimization is required to locate them. Coarser-grained mediators are more likely to optimize generality and are generally easier to search over, but tend to sacrifice selectivity and sparsity.

\subsection{Full layers and submodules}\label{ssec:layers-submodules}
Full layers $\mathbf{h^\ell}$ and submodules $\mathbf{h}^{\ell\text{-MLP}}$, $\mathbf{h}^{\ell\text{-Attn}}$ are relatively coarse-grained mediators; there exist only $\ell$ of them in a model $\mathcal{C}$.\footnote{Henceforth, we will refer to each of these with $\mathbf{h^\ell}$ for concision, as they are equivalent with respect to what kinds of search methods can be applied to them.}
Early probing classifiers studied the information encoded in full layers \citep{shi-etal-2016-string,hupkes2018visualisation,belinkov-etal-2017-neural,conneau-etal-2018-cram,liu-etal-2019-linguistic,hewitt-manning-2019-structural,giulianelli-etal-2018-hood}, 
and recent studies that leverage classifiers as part of causal techniques still frequently do the same \citep[e.g.,][]{elazar-etal-2021-amnesic,marks2023geometry,li2023emergent}.
This makes layers a natural mediator for exploratory interventions where using more fine-grained mediators is infeasible, as in \citet{conmy2023automated}, or where broad characterizations of information flow are sufficient, as in \citet{geva2023dissecting, sensharma2024locating}.

Full layers are rare mediators in mechanistic interpretability. This is because $\mathbf{h^\ell}$ is a \textbf{causal bottleneck}, such that all information in the model must pass through it. For example, if we run input $x_i$ through a model, and intervene on $\mathbf{h^\ell}$ to set it to what it would have been given another input $x_i'$, the model's output would be identical to the case where we simply change the input to $x_i'$ and perform no interventions. That said, interventions to full layers were employed in a pruning study where the motivation was not interpretability \citep{sajjad2023effect}; we believe that a more refined intervention technique has the potential to inform our understanding of which model \emph{regions} (as opposed to components) are more responsible for certain  behaviors~\citep[e.g.][]{lad2024remarkablerobustnessllmsstages}.\footnote{Relatedly, one could intervene on submodules, such as $\mathbf{h}^{\ell\text{-MLP}}$, and observe how this impacts the model, or the features present in the final layer output $\mathbf{h}^\ell$.} Past work has used coarse-grained methods based on full layers to investigate factual recall in language models \citep{geva2023dissecting}, and to update these factual associations \citep{meng2022locating,meng2023memit}.

The primary advantage of using full layers and submodules as mediators is their small quantity and broad scope of information (high generality). This means that even slow or resource-intensive methods will generally be easy to apply to all layers. In some cases, this is sufficient. However, an obvious disadvantage is that full layers are generally difficult to understand, as they are not particularly sparse nor selective: even if we know that a concept is encoded in a given layer, it is unclear precisely how this information is encoded, composed, or used, and how we might intervene on it without affecting other concepts \citep{conmy2023automated}. Thus, layers and submodules will generally be too coarse to explain model behavior or verify mechanistic hypotheses; under these goals, one is better off using full layers as tools to narrow the search space for harder-to-localize mediators~\citep[e.g.,][]{brinkmann2024mechanistic,geva2023dissecting}. However, if one's goal is localization or model editing, then full layers may be sufficient \citep{meng2022locating,meng2023memit,sensharma2024locating, gandikota2023erasing,gandikota2024unified}.

\subsection{Basis-aligned subspaces}
\label{sec:neurons_heads}
\paragraph{Neurons}
Compared to a full layer or submodule $\mathbf{h^\ell}$, a neuron $\mathbf{h}_i^\ell$ represents a more fine-grained component that could feasibly represent an individual concept (though we discuss below that this is not often the case due to polysemanticity). A neuron can be considered the smallest meaningful unit in the computation graph $\mathcal{C}$; the neuron's activation $h_i^\ell$ is a scalar corresponding to a single dimension (1-dimensional subspace) of a hidden representation vector.
Each neuron can sometimes be distinguished from others based on its functional role in the network;
for instance, \citet{BauBSDDG19} locate neurons in a machine translation model responsible for detecting or generating items of a particular tense, gender, or number, and causally verify their roles by intervening on their activations in a targeted manner.
\citet{bau2020understanding} locate neurons in a GAN responsible for generating specific types of objects in images, such as trees or windows, and verify this causally by ablating or artificially activating them.

Neurons are a natural choice of mediator, as they are both fine-grained (sparse) and easy to exhaustively iterate over (see \S\ref{ssec:exhaustive-search}); there are $O(\ell\cdot d)$ of them, where $d$ is size of the activation vector $|\mathbf{h}^\ell|$.\footnote{We use big-$O$ notation because the exact number will depend on whether one includes just the activation vectors at the end of a layer, or additionally includes the vectors output by the MLP and/or attention blocks, (among other possible vectors). Also, if looking at intermediate MLP neurons, the number is typically $e\cdot|\mathbf{h}^\ell|$, where $e$ is some constant expansion factor, typically in $[1.5, 4.0]$.}

However, a major disadvantage of neuron-based interpretability methods is \textit{polysemanticity}. 
Individual neurons are often polysemantic---i.e. they respond to multiple seemingly unrelated concepts simultaneously \citep{arora2018linear}, such that a neuron may be \emph{sparse} but not \emph{selective} nor \emph{human-understandable}. This gives them relatively low utility for explaining model behaviors, but they can be useful for verifying mechanistic hypotheses \citep{geiger2021causal} or localizing model behaviors \citep{vig2020causal}. 
For example, if the same neuron were sensitive to capitalized words, animal names, one-digit numbers, among other phenomena, and a researcher were to inspect the activations of that neuron, it would be difficult to disentangle each of these individual roles.
\citet{elhage2022superposition} investigate polysemanticity and suggest that neural networks represent features through linear superposition, where they represent features along non-basis-aligned linear subspaces, resulting in interpretable units being smeared across multiple neurons. 
In other words, in an activation vector $\mathbf{h}^\ell$ of size $|\mathbf{h}^\ell|=d$, a model can encode $k \gg d$ concepts as directions \citep{park2023linear}, such that only a sparse subset of concepts are active given a particular input. 

\paragraph{Basis-aligned multidimensional subspaces}
The computations of a neuron $\mathbf{h}_i^\ell$ are often not independent: \emph{sets of} neurons can compose to encode some concept. 
For example, in language models, subsets of neurons $\{\mathbf{h}_i^\ell, \mathbf{h}_j^\ell,\ldots\}\in \mathbf{h}^\ell$ can be implicated in
encoding gender bias \citep{vig2020causal}, and implementing latent linguistic phenomena \citep{finlayson-etal-2021-causal,mueller-etal-2022-causal,BauBSDDG19,lakretz-etal-2019-emergence}.
Thus, some early mechanistic interpretability work employed heuristic-based searches over sets of neuron responsible for some behavior \citep[e.g.,][]{bau2018visualizing,vig2020causal,cao-etal-2021-low,antverg2022on}. 
This is a generalization of individual neurons as mediators, where multiple dimensions in activation space are intervened upon simultaneously.

Sets of neurons have strictly more expressive power than individual neurons, and thus have the potential to explain model behavior more broadly than finer-grained mediators. In other words, one can conceptualize neuron groups as trading off some sparsity for increased generality. If a concept is encoded across multiple neurons, then neuron groups may also enable more human-interpretable interventions than one-dimensional subspaces.
Despite this, basis-aligned multidimensional subspaces are not commonly studied. 
This is for two main reasons: (1) There is a combinatorial explosion when we are allowed to search over arbitrarily-sized sets of neurons; if using exhaustive search, this increases the number of required forward passes from $O(\ell \cdot d)$ to $O(2^{\ell \cdot d})$, which makes this intractable. (2) Furthermore, interpretable concepts are not guaranteed to be aligned to neuron bases, meaning that groups of neurons still do not directly address the problem of polysemanticity~\citep{s.2018on,chugtai2023toy,wang2023interpretability}.

\paragraph{Attention heads}
Similar to neurons, attention heads are fundamental components of Transformer-based neural networks. They mediate the flow of information between token positions~\citep{vaswani2017attention}; thus, using attention heads as units of causal analysis can help us understand how models synthesize contextual information~\citep{ma-etal-2021-contributions,neo2024interpreting} to predict subsequent tokens~\citep{wang2023interpretability,greater_than,prakash2024fine,garcíacarrasco2024doesgpt2predictacronyms, brinkmann2024mechanistic}.

Each head $A_i^\ell$ in layer $\ell$ can be understood as an independent operation contributing a vector output $\mathbf{a}_i^\ell$; the outputs of all heads are concatenated and projected to form the output of the attention block $\mathbf{h}^{\ell\text{-Attn}}$, which is then added into the residual stream $\mathbf{h}^\ell$.\footnote{This is the \emph{residual stream perspective} \citep{elhage2021mathematical} of Transformers, which has been adopted in recent interpretability research \citep{ferrando2024primer}. The \emph{residual stream perspective} suggests that the residual stream, which comprises the sum of the outputs of all the previous layers and the original input embedding, acts as a passive communication channel through which the MLP and attention submodules route the information they add.}
For example, some heads specialize on syntactic relationships~\citep{chen2024sudden}, others on semantic relationships such as co-reference \citep{vig2020causal}, and others still on maintaining long-range dependencies in text~\citep{wu2024retrieval}.
Attention heads have also been directly implicated in acquiring the ability to perform in-context learning~\citep{olsson2022context,brown-2020-gpt3}, or to detect and encode functions in latent space \citep{todd2024function,feucht2025dualroute}.

Attention heads are attractive mediators because they are easily enumerable: there are generally far fewer attention heads $O(\ell\cdot |A^\ell|)$than neurons $O(\ell\cdot d)$ in a model, as $|A^\ell|\ll d$ in typical Transformer-based models. Attention heads also track multi-token relationships. However, in contrast to the activation $h_i^\ell$ of a neuron, the output of an attention head $\mathbf{a}_i^\ell$ is multidimensional. Thus, it is difficult to directly interpret the full set of functional roles a single head might have: attention heads are almost always polysemantic, so one cannot typically determine the function(s) of an attention head solely by observing its outputs \citep{janiak2023polysemantic}.\footnote{However, there is initial evidence that some dimensions of an attention head's output can be meaningfully explained \citep{merullo2024circuit, merullo2024talkingheads, hu2025understanding}. Thus, by decomposing the vector output of a head into smaller subspaces or even individual neurons, it may be easier to explain the set of functional roles of a given head.}
It has additionally been observed that intervening on an attention head can cause other attention heads to compensate, which further complicates causal analyses \citep{jermyn2023attention,wang2023interpretability,mcgrath2023hydra}.\footnote{This phenomenon where downstream components only have causal relevance after an upstream component has been removed is sometimes called \textbf{preemption} in the causality literature \citep{mueller2024missedcausesambiguouseffects}, or the ``Hydra effect'' in mechanistic interpretability \citep{mcgrath2023hydra}. Preemption is not limited to attention heads; future work should analyze how common preemption is between other types of components, such as MLP submodules.} To summarize, attention heads are easier to exhaustively search over than (sets of) neurons, but have the same issues of low selectivity and human-interpretability.

\subsection{Non-basis-aligned spaces}\label{sec:non_basis_subspaces}
\paragraph{Non-basis-aligned multidimensional subspaces} Due to their polysemanticity, neurons, attention heads, and sets thereof do not necessarily correspond to cleanly interpretable concepts. For example, individual neurons typically activate on many seemingly unrelated inputs~\citep{elhage2022superposition}, and this issue cannot be cleanly resolved by adding more dimensions. This is because the features may actually be encoded in directions or subspaces that are \emph{not aligned to neuron bases} \citep{mikolov-2013-word2vec1, arora-etal-2016-latent};
\S\ref{ssec:continuous-search} defines and visualizes this concept in more detail.

To overcome this disadvantage, one can generalize causal mediators to include arbitrary \emph{non-basis-aligned}  subspaces of $\mathbf{h}^\ell$. This allows us to capture more sophisticated causal abstractions encoded in latent space, such as causal nodes corresponding to greater-than relationships \citep{wu2023interpretability}, or equality relationships \citep{geiger2024finding}.
Common methods for locating these are discussed in \S\ref{ssec:continuous-search}.

The primary advantage of considering an arbitrary subspace as a mediator is its expressivity \emph{and} precision: subspaces often capture distributed abstractions that are not cleanly aligned to specific neurons. However, they are generally more difficult to locate than basis-aligned components or lower-dimensional directions, as we are required to have specific hypotheses as to how a model accomplishes a task and access to minimal pairs that isolate the target concept. Some optimization-based procedure is also usually required. Thus, non-basis-aligned multidimensional subspaces are generally more selective and human-interpretable than basis-aligned subspaces, and can maintain similar sparsity and generality; however, they may sacrifice faithfulness, as the optimized component can in theory learn concepts that were not in the model itself. This combined with the strict data requirements makes them useful primarily for verifying mechanistic hypotheses, but less useful for explaining model behaviors or localization/editing.

\paragraph{Directions} A recent line of work aims to automatically identify specific directions (one-dimensional non-basis-aligned objects) $\mathbf{d}$ that correspond to monosemantic concept representations; see Eq.~\ref{eq:direction} for a definition of $\mathbf{d}$. Identifying and labeling these monosemantic model abstractions (often called \textbf{features}; \citealp{bricken2023monosemanticity,cunningham2024sparse, huang2024ravel}) can reveal units of computation the model uses to solve tasks in a way that is often easier for humans to interpret.\footnote{Note that these directions are not necessarily subspaces of activation space: there are often non-linearities used in computing them, even though the vectors in activation space are involved in computing the directions. Therefore, we will refer to any one-dimensional space as a \textbf{direction}, but do not require it to be a subspace of activation space.}

There is also initial evidence that these directions may enable fine-grained model control \citep{panickssery2024steeringllama2contrastive, marks2024sparse, tigges2023linear}. Past work has found initial signs that basis-aligned directions could be leveraged to edit \citep{meng2022locating} or steer \citep{subramani-etal-2022-extracting,turner2023activation} model behavior, whereas more recent work has found that steering using non-basis aligned directions is both more effective and more precise \citep{marks2024sparse,arora2024causalgym,wu2025axbenchsteeringllmssimple}. For example, there is work that uses linear probes to understand the effects of a direction on the model behavior \cite{chen2024designing,ravfogel-etal-2020-null,elazar-etal-2021-amnesic,ravfogel-etal-2021-counterfactual,lasri-etal-2022-probing,marks2023geometry}, as well as work that uses these directions to steer model behaviors---e.g., by minimizing \citep{marks2024sparse,cunningham2024sparse} or amplifying \citep{templeton2024scaling} directions corresponding to fine-grained concepts, such as typically female names or the Golden Gate Bridge.

Nonetheless, directions have key disadvantages. The search space over possible non-basis-aligned directions is infinite, making it impossible to exhaustively search over them. To discover them, we are generally \emph{required} to modify the computation graph in some way to obtain some discrete search space---for example, by learning coefficients on each neuron, as in Eq.~\ref{eq:direction}, requires learning new parameters. Regardless of the method, optimization algorithms will introduce confounds due to their stochastic nature. In short, non-basis-aligned directions have significant advantages in human-interpretability, selectivity, and sparsity over basis-aligned mediator types, and their data requirements are less strict than non-basis-aligned multidimensional subspaces. This makes them a good starting point if one's goal is to explain model behaviors. Nonetheless, faithfulness is likely to be worse than that of basis-aligned components, given that optimization is required; generality may also suffer because of reconstruction error, and because the mediators are so fine-grained.

\subsection{Non-linear Mediators}\label{ssec:non-linear}
Non-basis-aligned directions and subspaces can be the most sparse and selective \emph{linear} mediator types. However, recent work has demonstrated that some features in language models can be represented non-linearly.
For example, there exist
features that are encoded as vector \emph{magnitudes} in any direction \citep{csordas-etal-2024-recurrent}. Past work has similarly found that many concepts can be more easily extracted using non-linear probes \citep{liu-etal-2019-linguistic}, and that non-linear concept erasure techniques tend to outperform strictly linear techniques \citep{iskander-etal-2023-shielded,ravfogel-etal-2022-adversarial}. 
However, in causal and mechanistic interpretability, most work has thus far tended toward using linear representations as units of causal analysis. Thus, there is significant potential in future work for systematically locating non-linearly-represented features---e.g., using group sparse autoencoders~\citep{10095958}, which could isolate multiple directions simultaneously, and/or probing and clustering techniques to identify multidimensional features \citep{engels2024not}. Non-linear features have not been extensively studied, despite their expressivity; we therefore advocate investigating these mediators in \S\ref{sec:discussion}.

\section{Searching for task-relevant mediators}\label{sec:mediator_search}
Once one has selected a task and a type of mediator, how does one identify task-relevant mediators of that type? The answer depends largely on the type of mediator chosen. If a given mediator type is finite in number---as is the case for sets of model components $\mathbf{h}^\ell$ such as neurons, layers, and submodules---one could perform an exhaustive search over all possible mediators, choosing which to keep according to some metric; \S\ref{ssec:exhaustive-search} discusses this approach. However, other mediator types, including non-basis-aligned directions and subspaces, are continuous, rendering an exhaustive search impossible. There are two common solutions to the problem of continuous mediator search spaces: (i) employ optimization to search this space, or (ii) narrow the space into an enumerable discrete set. Both are discussed in \S\ref{ssec:continuous-search}. Table~\ref{tab:mediators} (\S\ref{sec:mediator_type}) summarizes the kinds of mediators that tend to be paired with particular search methods. We do not discuss runtimes in this section, but see Appendix~B for an overview of the relative computational requirements for each mediator/search method combination.

\subsection{Exhaustive search}\label{ssec:exhaustive-search}
Suppose we are given a neural network with a finite set of candidate mediators $\{Z_i\}_{i=1}^N$, such as the set of all neurons $\{\mathbf{h}_0^0,\ldots,\mathbf{h}_d^\ell\}$. One way to identify task-relevant mediators from this set is to assign each mediator $Z_i$ a task-relevancy score $S(Z_i)$ and then select the mediators with the top scores. $S$ is typically the indirect effect (IE; \citealp{pearl2001effects,robins1992indirect}), as defined in Eq.~\ref{eq:ie}.\footnote{Other causal metrics include the \textbf{direct effect}, which measures the direct influence of the input on the output behavior except via the mediator. While more rarely used, it can be a helpful metric in tandem with indirect effects, as in \citet{vig2020causal}. There is also the \textbf{total effect}, which is the impact of changing the input on the model's output behavior. Note that the total effect does not directly implicate any particular component in model behavior, as it depends only on the input.} Computing this generally entails iterating over each mediator $Z_i$, setting its activation to some counterfactual value (either from a different input where the answer is flipped, or a value that destroys the information in the neuron, such as its mean value), and measuring how much this intervention changes the output. For example, \citet{vig2020causal} and \citet{finlayson-etal-2021-causal} perform counterfactual interventions to the activation of each neuron in an LM, measuring how much each intervention changes the probability of correct completions. This metric is based on the notion of counterfactual dependence, where we measure the difference in some output metric $m$ before and after intervening on a given component $Z_i$.

Exhaustive searches have many advantages: their results are comprehensive, causally efficacious, and relatively conceptually precise if our mediators are fine-grained units like neurons. We are also not required to have a pre-existing causal hypothesis as to how or where in its computations a model performs a task: we may simply observe how interventions to a model component changes the model's behavior or the probability of some continuation. Because of these advantages, this method is common when we have a finite set of mediators---for example, in neuron-based analyses \citep{vig2020causal,geiger2021causal,finlayson-etal-2021-causal} or attention-head-based analyses \citep{vig2020causal,conmy2023automated,syed2023attribution}.

However, exhaustive searches also have two significant disadvantages. The most obvious is that, in exact form, an exhaustive search requires $O(N)$ forward passes, where $N$ is the number of mediators. This does not scale efficiently as models scale, both because the number of components increases and because the computational cost of inference scales with model size.
This may be why exhaustive searches have not often been extended to \emph{sets of} neurons or heads, as this results in a combinatorial explosion in the size of the search space such that the number of required forward passes increases to $O(2^N)$.
Searches over sets of components can be approximated using greedy or top-k approaches, as in \citet{vig2020causal}, but this is not guaranteed to find the best solution to the problem of assigning causal credit to groups of components. 

To overcome these challenge, gradient attributions have become common. These can be conceptualized as fast linear approximations to causal mediation analysis. As they are local linear approximations, gradient attributions are technically not causal\footnote{Gradient attributions provide a scalar value whose magnitude can be interpreted as a local linear approximation of the model output's sensitivity to the component. This is conceptually related but distinct from causal analysis---e.g., because it is sensitive to confounding, does not capture non-linear effects, and does not directly test a causal model nor counterfactual hypothesis.} and not always accurate, but they have far better asymptotic runtime than causal mediation analysis. Example methods include attribution patching \citep{kramar2024atp,syed2023attribution}, and improved versions thereof inspired by integrated gradients \citep{sundararajan2017ig,marks2024sparse,hanna2024faithfaithfulnessgoingcircuit}. These techniques usually entail backpropagating from some target metric $m$; this is typically the probability of the correct next token or correct label $y$, or the probability difference between $y$ and some minimally differing incorrect output $y'$. This yields $\frac{\partial m}{\partial h^\ell_i}\Big|_x$ (the gradient of $m$ with respect to the activation of neuron $h^\ell_i$ given input $x$), which can be conceptualized as a local estimate of the slope of $m$ with respect to $h^\ell_i$. If we multiply this slope by the difference in $h_i^\ell$ and a counterfactual value ${h^\ell_i}'$,\footnote{A common counterfactual value includes $\mathbb{E}_{x\in X}[h_i^\ell]$; setting $h_i^\ell$ to this mean is known as a mean ablation. Setting $h_i^\ell$ to 0 is known as a zero ablation. See Appendix~A for more details.} then we can obtain a linear approximation of how much changing $h^\ell_i$ would have changed $m$---in other words, a linear approximation of the indirect effect (Eq.~\ref{eq:ie}). We can perform this attribution for all $h^\ell_i$ in parallel using $O(1)$ forward and backward passes.

The second and more difficult disadvantage to overcome is that exhaustive search constrains us to finite sets of mediators. Thus, this approach will not be possible if the search space is continuous (infinitely large). This is a key motivation behind the methods in the following subsection.

\begin{figure}
    \centering
    \includegraphics[width=0.8\linewidth]{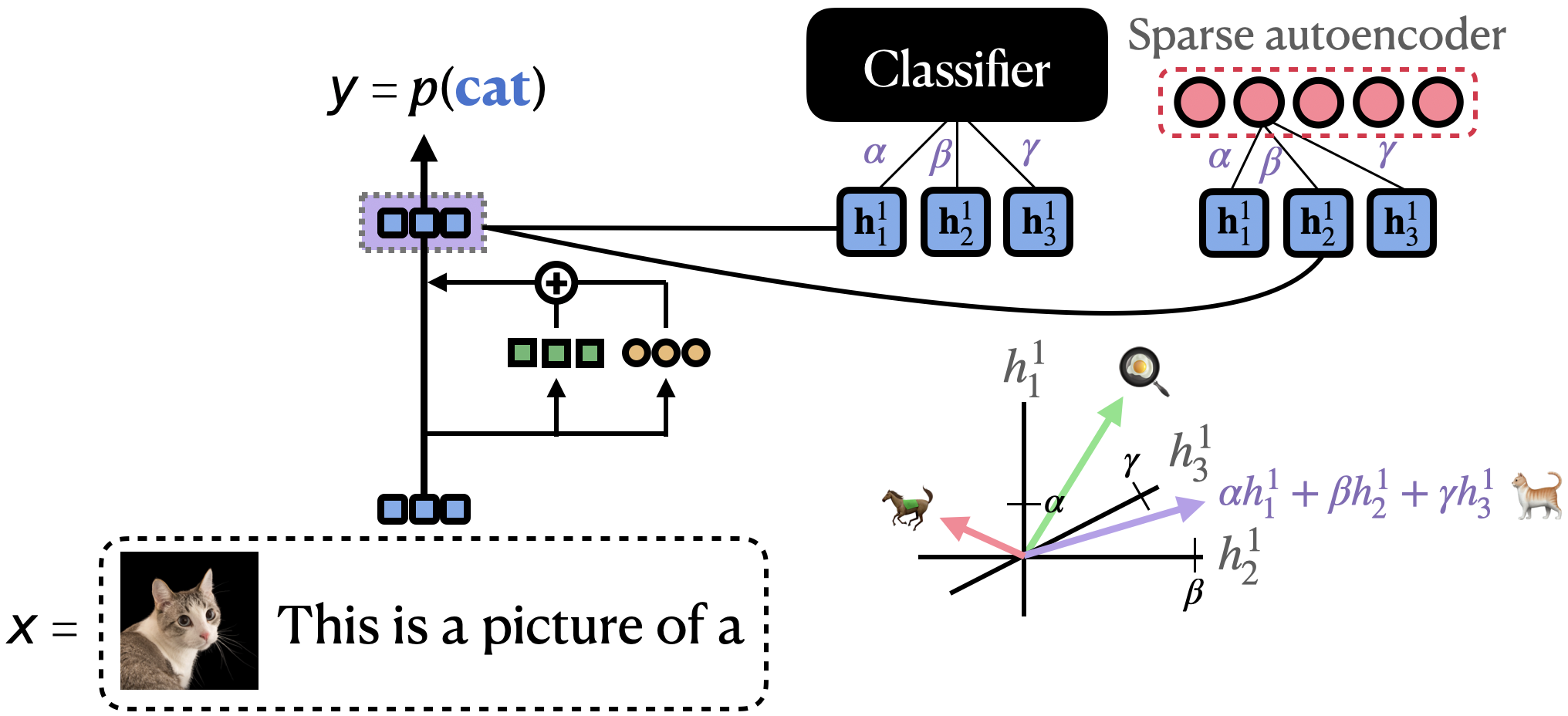}
    \caption{Neurons are not guaranteed to encode interpretable features. If non-basis-aligned directions encode the features of interest, then a neuron may activate on many different features that are non-orthogonal to its basis. Locating non-basis-aligned mediators requires components in addition to the model's computation graph that encode the coefficients on each activation. One can, for example, obtain these coefficients via supervised optimization with probing classifiers (\S\ref{sssec:supervised}) or unsupervised optimization with sparse autoencoders (\S\ref{sssec:unsupervised}). Note that optimization-based techniques sometimes introduce non-linearities, meaning that the discovered directions will not necessarily be a subspace of activation space.}
    \label{fig:non-basis-aligned}
\end{figure}

\subsection{Optimizing over large spaces of mediators}\label{ssec:continuous-search}
For some types of mediators, the collection of candidate mediators is continuous or far too large to exhaustively search over; this precludes using methods described in \S\ref{ssec:exhaustive-search}. To search over large but enumerable sets, some researchers employ modified versions of exhaustive search, including greedy search methods \citep{vig2020causal} or manual searches \citep{wang2023interpretability}. For continuous spaces, however, interpretability researchers generally use optimization. We taxonomize these optimization problems based on whether they require the interpretability researcher to manually select and incorporate task-specific information into the loss function (supervised methods; \S\ref{sssec:supervised}) or not (unsupervised methods; \S\ref{sssec:unsupervised}). We illustrate the intuition behind optimization-based search in Figure~\ref{fig:non-basis-aligned}.

\subsubsection{Supervised mediator searches}\label{sssec:supervised}
By \textit{supervised mediator searches}, we mean parametric approaches that require labeled task data $\mathcal{D}$ and/or human-provided hypothesized high-level causal graphs $\mathcal{H}$. For example, these methods might require the researcher to propose candidate intermediate concepts $V_i\in V$ that they expect the model to use in performing some task, or a candidate mechanism (a subset of $\mathcal{H}$) by which the model might complete the task. Others might simply require labeled data $(x,y)\in\mathcal{D}$ for training classifiers.

\paragraph{Supervised probing}
In supervised probing approaches, the researcher hypothesizes that the model represents some concept $V_i\in V$, designs some labeled dataset consisting of \mbox{$\{(x_i,y_i)\}_{i=1}^n\in \mathcal{D}$} to isolate the concept, and trains a probe to locate some direction(s) in $Z$ that \textit{correlate(s)} with the concept. A probe is generally formulated as a classifier $\Pi$ that maps from model representations $Z$ given input $x_i$ to probabilities over class predictions: $p_\Pi(y_i\mid x_i,Z)$.
Generally, $Z$ is a representation vector $\mathbf{h}^\ell$ at the end of a particular layer $\ell$, and the probe updates its weights over all possible subspaces therein to attend to  signals that are predictive of the labels. 
Many papers employ probing classifiers \citep{belinkov-glass-2019-analysis}, though most do not validate the causal efficacy of the probe \citep{belinkov2021ProbingSurvey}. A drawback of this is that certain activation subspaces are often \textit{correlated} with a concept without \textit{causally mediating} the concept. This means that probing can return many false positives \citep{hewitt-liang-2019-designing,elazar-etal-2021-amnesic,ravichander-etal-2021-probing,amini-etal-2023-naturalistic,belinkov2021ProbingSurvey}.

Thus, much recent work complements supervised probing approaches with additional checks of causality. For example, one can apply causal mediation analysis to the directions $W^\top_\Pi\mathbf{h}^\ell$ identified by supervised probing \citep{marks2023geometry, nanda-etal-2023-emergent}---i.e., one can measure whether the IE (Eq.~\ref{eq:ie}) of the direction used by the probe is significantly different from zero. This would fall under non-basis-aligned mediators. One can also backpropagate from the classifier to modify the behavior of the model \citep{giulianelli-etal-2018-hood} or to generate counterfactual representations \citep{tucker-etal-2021-modified}, which is more akin to a full-layer intervention. One can also directly compare the probe's predictions to a causally grounded probe \citep{amini-etal-2023-naturalistic}. 

Another line of work uses the directions discovered by probes to \emph{guard} or \emph{erase} information about a particular concept from the model's representations. For example, a direction in a model's activation space that is most predictive of the target concept can be nullified via orthogonal projections, such that the model can no longer use the information \citep{ravfogel-etal-2020-null}; this process can be repeated until linear guarding is achieved. The aim of this procedure is to make the indirect effect of the concept on the model's behavior equal to zero for all inputs.

Concept erasure and guarding can be used to measure the causal importance of particular concepts, as in \citet{elazar-etal-2021-amnesic},
though studies that employ methods like these tend to focus on single layers. More recently, techniques such as LEACE \citep{belrose2023leace} and follow-ups \citep{singh2024representationsurgerytheorypractice} have generalized this idea to provably prevent any linear classifier from using a concept; this moves beyond orthogonal projections and projects out the information at \emph{every} layer. One could use such methods to causally understand the \emph{set of} directions, or non-basis-aligned multidimensional space, that encode some concept. Note that many of these methods are still susceptible to the problems entailed by using linear mediators; thus, future work could follow \citet{iskander-etal-2023-shielded,ravfogel-etal-2022-adversarial} in generalizing these analyses to non-linear mediator types.

\paragraph{Counterfactual-based optimization} A related line of methods involves learning binary masks over sets of components (e.g. sets of neurons, attention heads, layers) to determine which are relevant mediators for a task. Examples include subnetwork probing \citep{cao-etal-2021-low} and Desiderata-based Component Masking (DCM) \citep{davies2023discovering, prakash2024fine}. These allow us to find basis-aligned multidimensional subspaces (\S\ref{sec:neurons_heads}), but they require triplets $(x_i,x_i',y_i)$ consisting of contrastive input pairs $(x_i,x_i')$ and a target $y_i$.

A more expressive class of approaches involves using the result of causal mediation analysis as a metric to directly optimize. One such line of methods includes Distributed Alignment Search (DAS) and follow-up methods such as Boundless DAS \citep{geiger2024finding, wu2023interpretability, huang2024ravel}. These methods are powerful in that they allow us to locate non-basis-aligned multidimensional subspaces (\S\ref{sec:non_basis_subspaces}) that correspond to a particular concept, though they have relatively strict data requirements compared to probing: we need not only labeled examples, but also contrastive input pairs \emph{and} a human-provided high-level causal graph. More precisely, given an activation vector $\mathbf{h}^\ell$ and hypothesized causal graph $\mathcal{H}$ containing nodes and edges $(V,E)$, the goal is to learn an invertible linear transformation $\mathbf{R}$ such that a concept of interest $V_i\in V$ is aligned to the bases of the transformed space $\mathbf{R}(\mathbf{h}^\ell)$. To locate the concept, one iterates over contrastive input pairs $(x_i,x_i')$ in training dataset $\mathcal{D}$ that vary only with respect to $V_i$ and (ideally) no other concepts. For each pair, interventions are performed in the transformed space such that $v_i$ (the value of $V_i$ given $x$) is set to a counterfactual value $v_i'$ that it would have taken given $x_i'$. Finally, the space is transformed back to the original space via $\mathbf{R}^{-1}$. This should result in a predictable change in model behavior, as defined by $\mathcal{H}$. See Appendix~C for an illustration and more detailed description. 

These methods provide time-efficient ways to search for human-interpretable variables encoded in intractably large or innumerable mediator sets. However, their key limitation is that they require either labeled data, contrastive pairs of inputs, and/or a pre-existing causal hypothesis as to how a model accomplishes some behavior. These methods can be evaluated with respect to accuracy in capturing model behavior, but they do not directly indicate \emph{a priori} what those hypotheses should be. When one obtains negative results, these methods also do not indicate in what specific ways the hypotheses are wrong. They also require sufficient training data to recover the concept of interest during training. As with all parametric methods, the above approaches are subject to overfitting or underfitting.

\subsubsection{Unsupervised mediator searches}
\label{sssec:unsupervised}

Supervised search methods (see \S\ref{sssec:supervised}) require specific hypotheses about the internal representations of neural networks. 
However, neural networks implement various behaviors, many of which may be counterintuitive to humans and therefore more likely to be missed in supervised settings.
For example, while \citet{li2023emergent} hypothesized a constant board state representation in a Transformer learning to play Othello, \citet{nanda-etal-2023-emergent} later found that the model actually switches the board state representation with every turn, taking the view of ``my pieces vs.\ opponent's pieces'' rather than ``black pieces vs.\ white pieces''.
This example demonstrates that it can be desirable to search for mediators without specifying beliefs as to what those mediators do ahead of time.

Hence, some studies employ \textit{unsupervised} methods. Typically, unsupervised methods return large---but finite---collections of mediators. Unsupervised methods are largely \emph{correlative}, meaning that the discovered mediators may not necessarily capture causally relevant or faithful aspects of a model's computation. However, the discovered mediators can then be implicated in a model's computation post hoc by
employing additional techniques, such as exhaustive searches for the highest-indirect-effect mediators (\S\ref{ssec:exhaustive-search}), to select task-relevant mediators from this collection.

\paragraph{Dictionary learning using sparse autoencoders} 
Exhaustive search for meaningful non-basis-aligned directions is impossible due to the infinite search space.
The \textit{dictionary learning} literature tackles this problem by performing an unsupervised search for directions in neuron activations which both (1) capture the information encoded in the internal representations and (2) are \textit{disentangled} from other meaningful directions. \citet{bengio2013representation} characterize disentangled representations as \textit{factors of variation} in the training dataset.

To identify these factors of variation, \citet{sharkey2023technical}~used sparse autoencoders (SAEs) to perform dictionary learning on a one-layer transformer, identifying a large (overcomplete) basis of features. 
Given activation vector $\mathbf{h}^\ell$, SAEs are trained to reconstruct $\mathbf{h}^\ell$ as $\hat{\mathbf{h}}^\ell$ while only activating a sparse subset of dictionary features. Concretely, SAEs typically consist of an encoder and decoder:
\begin{align}
    \mathbf{f} &= \text{ReLU}(W_e(\mathbf{h}^\ell - \mathbf{b}_d)+\mathbf{b}_e)\\
    \hat{\mathbf{h}}^\ell &= W_d\mathbf{f} + \mathbf{b}_d,
\end{align}
where $\mathbf{f}$ is the feature vector (the encoded space), $W$ denotes a learned weight matrix, and $\mathbf{b}$ denotes a learned bias vector.\footnote{This is a basic formulation of SAEs. There are more sophisticated architectures that have empirically demonstrated better reconstruction performance and/or more interpretable features \citep[e.g.,][]{rajamanoharan2024improvingdictionarylearninggated,braun2024identifyingfunctionallyimportantfeatures}.} It is common to refer to a single dimension of $\mathbf{f}$, or $\mathbf{f}_i$, as a \emph{feature}. \citet{cunningham2024sparse} applied SAEs to language models and demonstrated that the observed dictionary features are highly interpretable and can be used to localize and edit model behavior.
Since then, numerous researchers have found promising results in identifying funtionally relevant and human-interpretable features~\citep[][\textit{inter alia}]{templeton2024scaling, rajamanoharan2024improvingdictionarylearninggated, braun2024identifyingfunctionallyimportantfeatures, bricken2023monosemanticity,fel2025archetypalsaeadaptivestable}, many of them having predictable effects on the model's behavior under interventions. That said, SAEs are not able to perfectly reconstruct the model's activations, and may not be optimal for counterfactual operations such as steering \cite{wu2025axbenchsteeringllmssimple}.
Most importantly, however, we do not know \emph{a priori} what the ground truth features are in the model's computation, and can only use the reconstruction performance as a proxy measure of performance.

\paragraph{Correlation-based clustering}
Another unsupervised way of discovering meaningful units is clustering mediators by the similarity of their behavior over some dataset $\mathcal{D}$. This idea is not new \citep[cf.][]{Elman1990FindingSI}, but running causal verifications of the qualitative insights from clustering studies is relatively rare. 
\citet{BauBSDDG19} showed that neurons sharing a similar behavior are causally important for various functionalities in recurrent neural machine translation models.  
\citet{dalvi-etal-2020-analyzing} cluster neurons in language models, and are able to maintain performance after ablating a significant portion of them. The goal of \citeauthor{dalvi-etal-2020-analyzing}'s study was not interpretability, but their results nonetheless causally verify that redundancy is very common in neural networks.

There has recently been renewed interest in  mediator search via clustering. \citet{michaud2023the} propose a method to identify interpretable behaviors within neural networks by clustering parameters. Because the identified behaviors tend to be coherent, the units implicated in each cluster can be viewed as a set of components that have a functionally coherent role in the network. \citet{marks2024sparse} and \citet{engels2024not} generalize this from gradients to neuron or sparse autoencoder activations.
The activations that compose the clusters are then labeled according to the dataset samples on which they activate most highly. 
When based on neurons, clusters are basis-aligned subspaces; when based on sparse autoencoder features (as in \citealp{marks2024sparse}), they are non-basis-aligned subspaces. To compute the indirect effect of a cluster, one can intervene on each component in the cluster simultaneously and compute the resulting indirect effect. Then, to find the most causally relevant clusters, one can exhaustively search by taking the top cluster by indirect effect. This method has not yet been extensively employed or explored. However, intervening on elements within these clusters could be a useful way to establish the functional role of \emph{groups of} components in future work, or assess whether a subset of a model's behavior is implicated in a more complex task. We discuss this in \S\ref{sec:discussion}.

In summary, unsupervised mediator searches enable us to locate human-interpretable concepts from neural networks internals without any labeled data. They allow us to do so relatively quickly, and these concepts can then be causally implicated in model behavior post hoc. However, the recovered concepts are not guaranteed to be faithful to the model's true concept space, and we are never guaranteed to recover a complete nor non-redundant concept set. Another pressing issue in unsupervised search is evaluation: while it is possible to compare the relative quality of two different unsupervised interpretability searches, it is difficult to devise absolute metrics that meaningfully capture closeness to some ideal solution; indeed, an ideal or identifiable solution may not exist \citep{meloux2025everything}. For SAEs, some have tried to devise thorough evaluations \cite{karvonen2025saebenchcomprehensivebenchmarksparse}, but robust evaluations that enable comparisons between supervised and unsupervised methods have only just started to become common \citep[e.g.,][]{wu2025axbenchsteeringllmssimple,mueller2025mibmechanisticinterpretabilitybenchmark}.

\section{Discussion}\label{sec:discussion}
What is the right unit of analysis for describing the inner workings of neural networks? Thus far, we have categorized past work by mediator type and search method. Now, we turn our attention to practical questions: What kinds of studies can more easily be done with certain mediator types? Where is there still room to deploy less common mediators in useful ways? Our main goal is to encourage researchers to conceptualize the field in this way, in the hope that we can encourage more work on discovering better causal abstractions and better terms for discussing the inner workings of neural machine learning systems---and, more ambitiously, more rigorous theoretical foundations for interpretability.

\subsection{What is the right mediator?}\label{sec:right_mediator}
There are pros and cons to any mediator, and the best mediator will therefore depend on one's goals. In this section, we ask: When is it appropriate to deploy particular kinds of mediators and search methods? To answer this question, we revisit the three goals laid out in \S\ref{ssec:criteria} and the pros and cons discussed in \S\ref{sec:mediator_type} and \S\ref{sec:mediator_search}. We first hypothesize which mediators and search methods are likely to be best given recent evidence. Then, in the following subsections, we describe directions for future work that can further improve on these criteria, and call for work formalizing these criteria into concrete benchmarks. In Appendix~D, we provide an even more practical guide by giving concrete examples in specific research scenarios.

\paragraph{Explaining model behavior} If we wish to understand at an algorithmic level how a model performs some behavior, then in the absence of compute restrictions and with no strong prior hypotheses, \textbf{unsupervised optimization-based methods} (\S\ref{sssec:unsupervised}) over fine-grained mediators (such as \textbf{non-basis-aligned directions}, \S\ref{sec:non_basis_subspaces}) provide a strong starting point. For example, unsupervised methods like sparse autoencoders provide a fine-grained (selective) and human-interpretable interface to a model's computation, making explanations easier to derive in the absence of any pre-existing mechanistic hypotheses. They have also been found to yield higher faithfulness with fewer components compared to components like neurons \citep{marks2024sparse}, meaning that non-basis-aligned directions generally achieve a better trade-off between sparsity and faithfulness than basis-aligned directions. That said, autoencoder features are not guaranteed to be faithful to a neural network's behavior in next-token prediction, and they require either a human or an LLM to label or interpret the features, which is laborious and expensive~\citep{bills2023language, paulo2024automatically}. Moreover, natural language explanations of model components have inherent flaws \citep{huang-etal-2023-rigorously}: they may often exhibit both low precision and recall. Finally, non-basis-aligned directions, while more interpretable than basis-aligned components, require more human effort and/or compute than basis-aligned components to locate, and one may need to rediscover these directions if model fine-tuning, adaptation, or editing is part of the study.\footnote{Though \citet{prakash2024fine} find that the same model components are implicated in an entity tracking task before and after fine-tuning.}

\paragraph{Verifying a mechanistic hypothesis} If we already have a hypothesis as to how a model performs some behavior and wish to measure the accuracy of our hypothesis, then \textbf{multidimensional non-basis-aligned subspaces} (\S\ref{sec:non_basis_subspaces}) may be the right mediators, and a reasonable corresponding search method would be \textbf{counterfactual-based optimization} (\S\ref{sssec:supervised}). One can automatically search for the subspaces which correspond to a particular node in one's hypothesized causal graph using alignment search methods  \citep{geiger2024finding,wu2023interpretability,sun2025hyperdas,geiger2021causal}. Alignment search entails learning a linear transformation to isolate some target concept; this allows us to locate distributed representations that act as independent causal variables in non-basis-aligned spaces. This is relatively scalable, and enables us to qualitatively understand intermediate model computations. The primary downside is that we must anticipate the mechanisms that models employ to perform a task, as demonstrated in Appendix~C; if we cannot anticipate them, then curating data and refining one's causal hypotheses may require significant human effort. If these are significant concerns, then probing may be a better (though correlative and less precise) option. These methods are subject to the same confounds as any other optimization-based technique: the module we are optimizing could directly learn the phenomenon, rather than extracting it from the model \citep{hewitt-liang-2019-designing,sutter2025nonlinearrepresentationdilemmacausal}. 

\paragraph{Localization and editing} If one's goal is to localize some phenomenon in a model, then \textbf{exhaustive searches} (\S\ref{ssec:exhaustive-search}) over \textbf{basis-aligned subspaces} (\S\ref{sec:neurons_heads}) or \textbf{full layers} (\S\ref{ssec:layers-submodules}) may be sufficient. There are many comprehensive causal techniques for locating these, including causal tracing \citep{meng2022locating} and activation patching \citep{vig2020causal}, as well as techniques for locating graphs of basis-aligned mediators, such as circuit discovery algorithms \citep{goldowskydill2023localizing,wang2023interpretability,conmy2023automated}. Some of these methods are slow in their exact form, but fast approximations exist to these causal metrics, including attribution patching \citep{syed2023attribution} and improved versions thereof \citep{kramar2024atp,hanna2024faithfaithfulnessgoingcircuit, marks2024sparse}. Even in the absence of a deep understanding of the role of these mediators, localization can be useful for downstream applications like model editing \citep{meng2022locating,meng2023memit}\footnote{Though \citet{hase2023does} find that causal localizations do not always reflect the optimal locations for editing models.} and model steering \citep{todd2024function,goyal2020cace}. That said, if meaningful features are not actually aligned with neurons/heads, then we are not guaranteed to get the best performance unless we use more fine-grained and selective mediators. Future work should analyze the performance of model editing and steering methods when using different kinds of mediators. For example, \citet{marks2024sparse} compare the efficacy of debiasing approaches based on ablating neurons versus non-basis-aligned directions discovered via sparse autoencoders; they find that ablating non-basis-aligned directions is significantly more effective. \citet{wu2025axbenchsteeringllmssimple}, however, find that supervised approaches for locating non-basis-aligned directions (e.g., difference-in-means, \citealp{marks2023geometry}) are significantly more effective than sparse autoencoders. These studies represent promising initial steps toward the kind of principled comparison between mediators that we advocate.

\subsection{Suggestions for Future Work}\label{sec:future_work}
By centering the mediator type and the criteria defined in \S\ref{ssec:criteria}, we can gain new insights into the kinds of research that will be necessary to advance mechanistic interpretability. Here, we discuss lines of work we believe will be fruitful.

\subsubsection{Finding better causal mediators} There are almost certainly better causal mediators that have not yet been explored. By ``better'', we mean achieving a better Pareto optimum between the criteria described in \S\ref{ssec:criteria} for at least one of the listed goals. Current work on improving mediators tends to focus on non-basis-aligned directions, such as sparse features or directions discovered from supervised probing on the activations of a single layer/submodule $\mathbf{h}^\ell$. One could consider pursuing coarser-grained mediators by discovering multi-layer \textbf{model regions} or \textbf{component sets} that accomplish a single behavior. Because these regions can cross layers, they would include non-linearities that allow them to represent more complex functions or concepts. We believe this would improve the generality of our findings, but not necessarily sparsity nor human-interpretability. Thus, we believe coarser-grained mediators will be better-suited to verifying mechanistic hypotheses or localization and editing.

\paragraph{Non-linear and multidimensional feature discovery}
As discussed in \S\ref{ssec:non-linear}, there is recent work demonstrating the existence of human-interpretable \emph{multidimensional} features. For example, days of the week are encoded circularly as a set of 7 directions in a two-dimensional subspace \citep{engels2024not}, and current methods cannot easily capture these multidimensional features. The ability to discover these features could greatly improve our understanding of the feature space of NNs, and thus our ability to systematically explain more of their behavior (i.e., increase generality) in a faithful and human-interpretable way. Group sparse autoencoders \citep{10095958} or clusters of autoencoder features could be a way to capture multidimensional non-basis-aligned features in an unsupervised manner, but despite promising initial evidence, empirical work has not yet demonstrated to what degree this will be effective for interpreting or controlling NNs. Additionally, many current mechanistic interpretability methods require us to define binary distinctions between correct and incorrect answers, whereas causal mediation analysis does not have any theoretical linearity, dimensionality, or Boolean restrictions; thus, interpretability methods will need to be extended to handle new kinds of variables.

There may also exist higher-order non-linear concepts in latent space that we have not yet located due to the linear focus of contemporary methods. For example, a subgraph or subcircuit can encode a coherent variable representation or functional role, as in \citet{lepori2023uncoveringintermediatevariablestransformers,li2024circuit}. How can we discover these subgraphs?
Path patching \citep{goldowskydill2023localizing,wang2023interpretability} provides a manual approach to implicating subgraphs as causal mediators, but we do not yet have automatic methods that can scalably search over subgraphs of a computation graph. Even if we were able to locate them, how might non-linear and/or coarse-grained mediators like these be useful in practice? As an example, we might expect fundamental phenomena like syntax processing to be spread across many layers of a language model. Syntax-sensitive components should be implicated in downstream tasks like question answering (QA) if we expect that language models are robustly parsing the meaning of the inputs. Thus, one could use causal mediation analysis to locate all components in the model with some high indirect effect on syntactic processing (e.g., using a subject--verb agreement task); this can be conceptualized as the syntax processing region of the model. Then, one could implicate the region(s) in the model's QA performance by intervening on each component in the region and observing whether performance changes. If the syntax processing region is not strongly implicated in QA performance, then we have a strong hint that the model may not be parsing the meaning of the inputs, but instead relying on a mixture of surface-level spurious heuristics---for example, memorized bigram associations, or giving answers with high prior probabilities. These examples illustrate how these more coarse-grained mediators could help us verify new kinds of mechanistic hypotheses.

\subsubsection{Inherently interpretable models} More ambitiously, one could consider building models with inherently interpretable components---i.e., whose fundamental units of computation (or some subset thereof) are designed to be sparse, monosemantic, and/or human-interpretable, but ideally still expressive enough to attain good performance on downstream tasks. Examples based in neural networks include differentiable masks \citep{de-cao-etal-2020-decisions,bastings-etal-2019-interpretable}, transcoders \citep{dunefsky2024transcodersinterpretablellmfeature}, codebook features \citep{tamkin2023codebookfeaturessparsediscrete}, and softmax linear units \citep{elhage2022solu}. These are primarily post hoc methods that decompose model components into interpretable units, but they could potentially be integrated into the network itself during pre-training alongside a loss term (in addition to a typical cross-entropy loss) that enables fine-grained interpretability at all stages of pre-training.

Alternatively, more focus could be devoted to building models that are designed from the ground up to be interpretable, such as backpack language models \citep{hewitt-etal-2023-backpack}, concept bottleneck models \citep{koh2020concept,oikarinen2023labelfree}, or decision trees \citep{hu2019optimal}. A related idea is to train the model using loss terms that encourage success on intermediate tasks, or induce particular kinds of feature representations \citep{hupkes2017diagnostic}. Perhaps least invasively, we could consider pre-training methods that softly encourage interpretable features to be aligned to neuron bases; this would remove the need for optimization to find non-basis-aligned components, and therefore make interpreting model decisions significantly easier and less confounded. However, this would reduce the number of features that could be encoded per neuron, so it would likely require significantly more parameters, or accepting degradations in performance. Regardless, we believe that this line of work will improve our ability to explain the behaviors of contemporary NN-based systems like neural language models via directly improving the human-understandability of their intermediate computations; this will make it far easier to explain model behaviors, verify mechanistic hypotheses, and localize/edit particular computations---but at the potential expense of performance on downstream tasks.

\subsubsection{Scalable search}
As the size of neural networks increases, the number of potential mediators to search over will also increase. The situation worsens as we start searching over continuous sets of fine-grained mediators such as non-basis-aligned directions. Although a few gradient-based or optimization-based approximations to causal influence have been proposed to improve time efficiency, such as attribution patching \citep{syed2023attribution} and DCM \citep{davies2023discovering}, more work is still needed to evaluate the efficacy of these techniques in identifying the correct causal mediators. Additionally, better techniques  beyond greedy search methods should be devised to identify causally important \emph{groups of} mediators; these should aim to produce Pareto improvements over time complexity and causal efficacy. 

As discussed in \S\ref{sssec:supervised}, optimization-based mediator search methods can be effective, but often require pre-existing hypotheses as to how a model implements a particular behavior. Thus, one could investigate using large language model agents to automate the process of hypothesis generation.
\citet{qiu2024phenomenal} showed that current LLMs can generate hypotheses, and \citet{shaham2024multimodal} showed that hypothesis refinement via LLMs can aid humans in interpreting the causal role of neurons in multimodal models. 
Similary, LLMs could be used to automate and scale hypothesis generation regarding the role of particular mediators across a wider variety of tasks and models.
Optimization-based based methods such as DAS or DCM could then be used to causally verify the automatically generated hypotheses, potentially in an iterative loop of hypothesis refinement and empirical testing.

\subsubsection{Benchmarking progress in mechanistic interpretability}
How will we know when we have made genuine improvements along any of the criteria that we have proposed? There exist few standardized methods or datasets for measuring general progress. To address this, research is needed on \textbf{standard benchmarks for measuring progress in mechanistic interpretability}. Currently, most studies develop ad hoc evaluations, and only compare to similar methods that employ the same mediators. Thus, to measure whether new mediators or search methods are truly giving us improvements over previous ones, we need to develop methods for performing principled direct comparisons. In circuit discovery, it is theoretically possible to use the same metrics to compare any circuit discovered for a particular model and task, regardless of whether sparse autoencoders are used, whether the circuit is based on nodes or edges, among other variations. Some recent work has begun to perform direct comparisons across mediator types, such as \citet{miller2024transformercircuitfaithfulnessmetrics}. \citet{huang2024ravel} propose to directly evaluate interpretability methods according to the generality of the abstractions they recover, and directly compare across different mediator types given the same model and task. \citet{arora2024causalgym} and \citet{makelov2024towards} also propose standardized interpretability benchmarks that allow us to compare across mediator search methods, though they do not directly compare across mediator types. 

Direct comparisons require defining criteria for success, but in mechanistic interpretability, there is little agreement about the kinds of phenomena we should be measuring and precisely how they should be measured (with the exception of faithfulness, which is very common but still not standardized; \citealp{hanna2024faithfaithfulnessgoingcircuit,wang2023interpretability}).  Some tasks have started to integrate human/user evaluations, which will be especially useful for building interpretability tools that are grounded in real-world use cases and settings \citep{saphra-etal-2024-first}. One benchmark that aims to enable direct comparison across mediator types and search methods is the Mechanistic Interpretability Benchmark (MIB; \citealp{mueller2025mibmechanisticinterpretabilitybenchmark}), which consists of two tracks: one for comparing circuits based on basis-aligned mediators, and another for comparing across mediator types/search methods. The fusion of these two tracks---i.e., comparing full causal graphs based on non-basis-aligned mediators, or even non-basis-aligned subspaces---could add value beyond the sum of these two separate tracks, and enable us to more directly benchmark progress on each of the goals of MI. Furthermore, \citeauthor{mueller2025mibmechanisticinterpretabilitybenchmark} only measure (counterfactual) faithfulness and sparsity, as these are the most tractable metrics at the relatively large scale needed for a benchmark. Future benchmarks should develop more scalable measures of generality and selectivity (e.g., via out-of-distribution evaluations).

For model editing and localization, more task-specific downstream metrics and datasets will be needed. For example, \citet{cohen2024evaluating} and \citet{zhong2023mquake} propose benchmarks to evaluate model editing methods on out-of-distribution examples, and \citet{karvonen2024measuring} propose to measure progress in feature disentanglement using board game models. \citet{wu2025axbenchsteeringllmssimple} define a measure for the quality of steering methods based on different mediator types and search methods. While not the main focus of this survey, we believe that building standardized benchmarks will be a key means to the end of assessing whether advancements in causal mediators are producing real improvements in applications of interpretability. More broadly, robust evaluation metrics and methods will lead to a more accurate science of the inner workings of language models, which will allow us to assess whether new causal abstractions are fundamentally more useful---for explaining the computations of a neural network, for verifying hypotheses, \emph{and} for practical applications.

\section{Conclusion}
In any study analyzing model behaviors via analyzing model components, the type of component(s) analyzed will determine what kinds of findings are possible. Some units are more closely aligned to the target concepts, while others are more faithful to the model's computation. Some units are easier to search over, but more difficult to understand (or vice versa). We have proposed a narrative and taxonomy of mechanistic interpretability research grounded in these units of analysis, or causal mediators. We have discussed the strengths and weaknesses of each mediator type, as well as what kinds of search methods are commonly employed for each. We have also discussed open problems in the field, focusing on those where this perspective reveals actionable and impactful research opportunities.

\appendix

\appendixsection{Types of interventions}\label{app:intervention-types}
To compute the indirect effect (Eq.~\ref{eq:ie}), we must replace the value $v_i$ of causal node $V_i$ with some counterfactual value $v_i'$. Say we are using mediator type $Z$, and that we are performing an exhaustive search by computing the IE for all $z_i\in Z$, and then taking the top components by IE. How should we compute $z_i'$? There are many ways to derive $z_i$: some of these depend on $x$, others depend on whether $x$ is a member of some class (e.g., inputs about dogs or inputs not about dogs), and others still depend on neither (e.g., are constant values or involve adding a noise term). Here, we briefly describe each of these classes of interventions, and describe how they will affect the kinds of components one will uncover.

Broadly, constant interventions will tell one which components have \emph{any} impact on model behavior, regardless of how. Input-dependent interventions are more precise, but tend to have lower recall: they isolate components whose impact changes when the input changes in the specific way defined by the intervention. Class-dependent interventions are a sort of medium between these.

\appendixsubsection{Input-dependent interventions}
\paragraph{Deterministic interventions}
If one cares about neurons that are sensitive to a specific contrast, then one can use input-dependent interventions (i.e., interventions where $z_i'$ depends on $x$). For example, assume our target task is subject--verb agreement. Given an input $x=$ ``The \textbf{\textcolor{blue}{key}}'', we want to locate neurons that increase the probability difference $m = p(\textbf{\textcolor{blue}{is}}) - p(\textbf{\textcolor{red}{are}})$. We obtain $z_i$ by running $x$ through model $\mathcal{C}$ in a forward pass (denoted $\mathcal{C}(x)$) and storing the activation $z_i$ of component(s) $Z_i$ (which could be a neuron, for example). We then obtain $z_i'$ by running $\mathcal{C}(x')$, where $x'$ is a minimally different input that swaps the answer; here, $x'$ would be ``The \textbf{\textcolor{red}{keys}}''. This type of intervention preserves example-specific information, and varies only the grammatical number of the subject. This will only reveal neurons for which swapping grammatical number \emph{and nothing else} will significantly affect the model's output.

Input-dependent interventions are precise: they reveal components targeted to a specific contrast between two prompts. However, humans must carefully curate controlled input pairs in which only one phenomenon is varied across $x$ and $x'$. Input-dependent interventions work best for \emph{binary} contrasts, where one defines two minimally different inputs that isolate neurons sensitive only to the difference between items in the pair. This yields counterfactuals that are semantically meaningful, making the results of an intervention easier to interpret. When working with categorical or ordinal variables, it is not immediately clear how to construct $x'$ to recover all relevant components. Additionally, it does not recover all task-relevant components; it only recovers those sensitive to the contrast between $x$ and $x'$. In other words, this is a low-recall method. For instance, \citet{vig2020causal} enumerates through all possible gender pronouns and nouns related to a specific gender to measure gender bias; they note that full generalizability to all grammatical gender pronouns is difficult. Furthermore, such interventions can be privy to unreliable explanations. This was shown in \citet{srivastava2023corrupting}, where the input data was corrupted to manipulate the concept assigned to a neuron. Hence, input-dependent interventions may require additional safeguards to ensure safety and fairness in critical real-life applications.

\paragraph{Stochastic interventions}
Another common intervention type entails adding noise to $z_i'$, without defining some specific $x_i'$ from which to derive it. For example, \citet{meng2022locating,meng2023memit} derive the counterfactual as $z_i' = z_i + \epsilon$, where $\epsilon\sim\mathcal{N}(0,3\sigma_{Z_i})$. $\sigma_{Z_i}$ is the standard deviation of $Z_i$ on some dataset. This intervention depends on $z_i$ by definition, but does not isolate a semantically meaningful contrast as input-dependent interventions do. This is conceptually closer to class- and independent-independent interventions (\S\ref{ssec:independent_interventions}), in that it will isolate components with \emph{any} impact on the model's behavior, regardless of the semantics of that impact. However, its stochasticity introduces variability in results and can be harder to interpret causally.

\appendixsubsection{Class-dependent interventions}
In contrast to input-dependent interventions, class-dependent interventions define a single intervention across a class of inputs. For example,~\citet{li2024circuit} learn a mask over the computational graph of a language model to prevent the model from producing toxic content; here, the two classes are \emph{toxic} and \emph{not toxic}, and the intervention within one of those classes is the same as for all other inputs in that class.

Class-dependent interventions provide a single flexible intervention that works for any given input. 
However, they require a dataset of input-label pairs that can be used to learn the interventions. 
This requires labeled data, and is sensitive to spurious correlations. Furthermore, many labels we care about are hard to definitively label without ambiguity (e.g., bias or toxicity).

\appendixsubsection{Class- and input-independent interventions}\label{ssec:independent_interventions} This type of intervention does not rely on the input nor a class label. The goal of these interventions is generally to fully remove (\emph{ablate}) the information encoded by a mediator, regardless of whether the information is task-relevant.\footnote{An ablation is a type of intervention. The goal of an ablation is to \emph{remove} the information stored in a component. ``Intervention'' is a broader term that refers to setting some $v_i$ to any value it would not naturally have taken.} A common ablation type is \textbf{zero ablations} \citep{dabkowski-2017-mean,lakretz-etal-2019-emergence,geva2023dissecting}, where the activation of a component is set to $0$. This is not entirely principled, since $0$ has no inherent meaning in an activation---for example, a neuron's default activation may be non-zero, whereas $0$ itself is out of distribution relative to what the model expects. A more principled ablation type is a \textbf{mean ablation} \citep{zeiler-2014-visualizing,ghorbani-2020-mean,mcdougall-etal-2024-copy}, where the neuron's activation is set to its mean value over some distribution---either task-specific data or general text data. A \textbf{resampling ablation} \citep{robnik-2008-resambpling,chan2022causal} is typically defined as a special case of mean ablations where the sample size is 1, and where the counterfactual input is randomly sampled.

Class- and input-independent interventions are a more general type of intervention that can be run without access to contrastive input/output pairs, and without labeled inputs. They allow us to tell whether \emph{any} of the information in a mediator is necessary for a model to perform the task, but they may also affect other information in unanticipated ways; in other words, they have high recall and low precision relative to the previous intervention types. They may also cause performance on a task to drop in a way that reveals spurious mediators, rather than mediators that are conceptually relevant. For example, in subject--verb agreement, ablating a neuron that detects the word ``dog'' may reduce the probability of the correct verb form ``is'' over the incorrect verb form ``are'', but this is a highly input-specific neuron that does not, in isolation, reveal general information about how models perform syntactic agreement.

\appendixsection{Computational Considerations}\label{app:compute}
We have briefly touched on the computational considerations inherent to each mediator type and search method throughout the survey. Here, we expand this discussion by more directly comparing their computational costs. 

\begin{table}[t]
    \centering
    \resizebox{\linewidth}{!}{
    \begin{tabular}{lcccc}
    \toprule
    Method & Layers/submodules & Neurons & Basis-aligned spaces & Non-basis-aligned spaces \\
    \midrule
    Exhaustive search & $O(\ell)$ & $O(\ell \cdot d)$ & $O(2^{\ell\cdot d})$ & N/A \\
    Gradient attribution & $O(1)$ & $O(1)$ & N/A$^*$ & N/A \\
    \midrule
    Probing & $O(\ell)$ & N/A & $O(\ell)$ & $O(\ell)$ \\
    Alignment search & $O(\ell)$ & $O(\ell \cdot d)^\dagger$ & $O(\ell)$ & $O(\ell)$ \\
    \midrule
    Sparse autoencoders & N/A & N/A & N/A & $O(\ell \cdot f)^\ddagger$ \\
    Clustering & $O(K)$ & $O(K)$ & $O(K)$ & $O(K)$ \\
    \bottomrule
    \end{tabular}}
    \caption{Summary of number of forward passes (and backward passes, when applicable) needed to locate the most causally relevant mediators of a given type (columns) using a particular method (rows). When a method involves training, we do not include training time in these estimates. $^*$One could operationalize this as the sum of neurons' gradient attributions (in which case it would be $O(1)$, though finding the best combination could still be exponential), but this is not recommended for three reasons: interaction effects, redundancy, and potential nonlinear compositions. $^\dagger$This estimate is based on the method of \citet{geiger2021causal}, but this is not common; with more recent methods like Boundless DAS \citep{wu2023interpretability}, it could in theory be reduced to $O(\ell)$. $^\ddagger$This assumes exhaustive search; the time becomes $O(1)$ if using gradient attributions.}
    \label{tab:compute}
\end{table}

Table~\ref{tab:compute} contains estimates of the number of forward and backward passes through $\mathcal{C}$ needed to locate the most causally relevant mediators, assuming we compute causal relevance using the indirect effect as in Eq.~\ref{eq:ie}. Where applicable, $\ell$ refers to the number of layers, $d$ to the size of an activation vector $\mathbf{h}^\ell$, $f$ to the size of an SAE's latent vector (i.e., the size of the output of the encoder), and $K$ to the number of clusters (a hyperparameter used in clustering algorithms). We see that gradient attributions are always fastest where applicable, but as they are linear estimates, we expect them to be less accurate than more exact methods like exhaustive search. 

Note that training times are excluded from these estimates. In general, training a single probe should have similar amortized runtime compared to a single training run of alignment search, though the difference lies in how many runs would be needed to find the correct features, and in how many examples would be needed to properly train them. If using a typical linear classification probe, one only needs to train $O(\ell)$ times maximum to obtain the best probe. If using Boundless DAS \citep{wu2023interpretability}, one needs to train $O(\ell \cdot |x|)$ times, where $|x|$ is the length of the input sequence. For unsupervised methods, training time can vary significantly; training a sparse autoencoder can take over a day given even a relatively small model of $<1\text{B}$ parameters (assuming access to one A100 GPU), but one only needs to train $O(\ell)$ of them. For clustering, the training time depends on the clustering method, but one can, in theory, cluster essentially any set of scalars relatively quickly. Moreover, clustering can be performed over all possible components in a model simultaneously without needing to iterate over layers (though it may sometimes be beneficial to perform clustering iteratively).

\appendixsection{Illustration of Alignment Search}\label{app:alignment-search}
\begin{figure}
    \centering
    \includegraphics[width=0.6\linewidth]{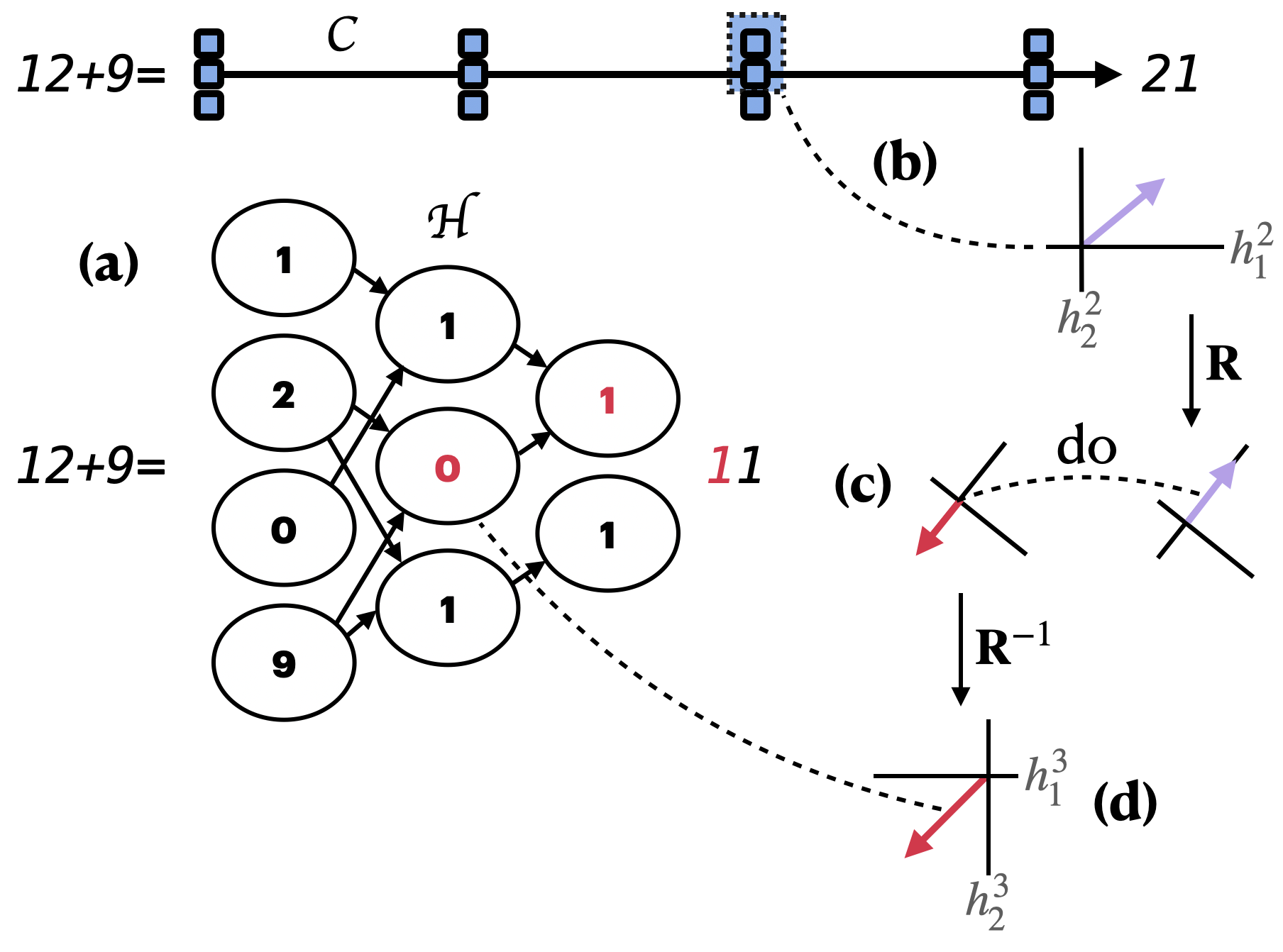}
    \caption{Example of alignment search, based on an example from \citet{mueller2025mibmechanisticinterpretabilitybenchmark}. (a) We start with the computation graph $\mathcal{C}$, and a hypothesized high-level causal graph $\mathcal{H}$. The hypothesis is that the model accomplishes addition using a tens-place addition, a ones-place addition, and a carry-the-one variable. (b) We hypothesize that the carry-the-one variable exists in layer two ($\mathbf{h}^2$). This variable may exist between multiple neurons, so interventions to neurons will not suffice. (c) We learn a rotation $\mathbf{R}$ into a new space where the target variable is aligned to the basis. This allows us to perform an intervention (the do-operation) to change the carry-the-one variable to some counterfactual value. (d) After intervening, we rotate back out using $\mathbf{R}^{-1}$. If the hypothesized causal graph is correct, the new output should be 11 instead of 21 after changing the carry-the-one variable's value.}
    \label{fig:alignment}
\end{figure}

Here, we provide an illustration of an alignment search example (Figure~\ref{fig:alignment}), as described in \S\ref{ssec:continuous-search}, and use this example to illustrate the goals of mechanistic interpretability more broadly. We have a hypothesis as to how the model $\mathcal{C}$ performs addition; in other words, we have a guess as to what $\mathcal{H}$ looks like. To isolate a variable in the hypothesized $\mathcal{H}$, we must design a dataset of contrastive pairs that vary only with respect to the variable. For example, if we believe the model contains a carry-the-one feature, we can design a dataset of inputs that vary with respect to whether the model must carry the one while leaving all other variables unchanged. We can use these pairs to isolate the variable during a training procedure; see \citet{wu2023interpretability,geiger2024finding} for more details.

Generally, one does not search for every variable in the hypothesized $\mathcal{H}$; one might select a few variables of interest. To quantify whether one has found them, one performs interventions to the discovered variable, as depicted in Figure~\ref{fig:alignment}. Success is measured by whether the predicted change to the model's output under the intervention given  $\mathcal{H}$ is what is actually observed.

\appendixsection{Concrete Examples}\label{app:examples}
The discussion in \S\ref{sec:right_mediator} is abstract. Here, we aim to give more concrete examples as to when certain kinds of mediators may be more appropriate.

Assume we are interested in understanding how a language model performs multiple-choice question answering. Also assume that we can afford to rerun fine-tuning and adaptation if this model is particularly bad, so we do not intend to perform precise model editing based on the results of benchmarking evaluations or interpretability experiments. Instead, we care mainly about predicting success and failure modes on future examples so that we know whether this model could be deployed in production, and in what cases we should double-check the model's outputs. In this case, the goal is to \emph{explain model behavior}, and we do not have a specific mechanistic hypothesis. Thus, we should deploy an \emph{unsupervised method} to locate and search over meaningful features, such as non-basis-aligned directions. This will help one find unantipicated mechanisms.

Assume instead that we want to precisely edit the knowledge of the model on cases where it gets the answer wrong. Now, we do not necessarily care as much about interpreting the model's general answering process (as helpful as this would be), but rather, debugging and fixing specific mistakes. Thus, this would fall under localization and editing: we would like to locate the source of incorrect answers, and patch them to improve performance. Thus, as a first step, we should deploy an exhaustive search over a relatively coarse-grained mediator, such as submodules or full layers. We can use model editing techniques like ROME \citep{meng2022locating} or MEMIT \citep{meng2023memit}, which perform targeted updates to basis-aligned components, to edit facts in cases where the model answered incorrectly. It is theoretically possible that localizing editing over non-basis-aligned mediators could result in even better performance. For example, AlphaEdit~\citep{fang2025alphaedit} performs a targeted update to the \emph{null space} of MLP layers, outperforming both ROME and MEMIT on several benchmarks. Future work should investigate whether this is possible, and whether the expected time complexity increase is worth the performance improvements.

Now assume that we are running a different kind of study: the task is still multiple-choice question answering, but we are testing a specific hypothesis as to how the model accomplishes the task. We want to know whether it represents ``truthfulness'' as an independent concept, and the study is only concerned with to what extent this holds---not accuracy on the task \emph{per se}. Here, we want to verify a specific mechanistic hypothesis, so we should design a dataset of labeled examples, where the label is based on truthfulness, and then deploy a supervised method such as probing over layers. If we can find a way to design counterfactual input pairs that vary only with respect to truthfulness, then we could instead deploy a more precise supervised method such as counterfactual-based optimization over non-basis-aligned subspaces. This would yield a set of scores that indicate to what extent the hypothesis causally explains the model's output behavior.

Note that none of these examples have recommended the use of basis-aligned subspaces, such as (sets of) neurons or attention heads. This is not to say that they are not useful, but it does indicate that when compute is not a significant limitation, they are often not the best place to start when working with realistic neural networks trained on large-scale data. Basis-aligned units are often difficult to interpret, and there are many of them; other mediator types are generally either more interpretable or easier to search over. That said, basis-aligned subspaces may be useful when we expect that they may have interpretable meanings (e.g., in toy task settings), or when we expect that unsupervised methods like sparse autoencoders are likely to yield bad results, or are simply not effectively trainable given one's resources.

\bibliographystyle{compling}
\bibliography{custom}

\end{document}